\documentclass[sn-mathphys]{sn-jnl}
\usepackage[utf8]{inputenc}

\usepackage{amsmath,amsfonts}
\usepackage{array}
\usepackage[caption=false,font=normalsize,labelfont=sf,textfont=sf]{subfig}
\usepackage{textcomp}
\usepackage{stfloats}
\usepackage{url}
\usepackage{verbatim}
\usepackage{balance}

\usepackage{lineno}
\usepackage{hyperref}
\usepackage{booktabs}
\usepackage{xcolor}
\usepackage{amsmath}
\usepackage{algpseudocode}
\usepackage{algorithm} 
\usepackage{multirow}
\usepackage{graphicx}
\usepackage[graphicx]{realboxes}
\usepackage{lscape}
\usepackage{float}
\usepackage{footmisc}
\usepackage{amsfonts}
\usepackage{stmaryrd}
\usepackage{booktabs}

\newfloat{algorithm}{t}{lop}
\DeclareMathOperator*{\argmax}{argmax}

\newcommand*\puboi[0]{PUBO$_i$}

\jyear{2025}
\begin{document}

\title[Neuro-LS]{Discovering new robust local search algorithms with neuro-evolution}

\author{\fnm{Mohamed Salim} \sur{Amri Sakhri}}\email{mohamedsalim.amrisakhri@univ-angers.fr}

\author{\fnm{Adrien} \sur{Go\"effon}}\email{adrien.goeffon@univ-angers.fr}

\author{\fnm{Olivier} \sur{Goudet}\footnote{Corresponding author. E-mail address: olivier.goudet@univ-angers.fr}}\email{olivier.goudet@univ-angers.fr}

\author{\fnm{Fr\'ed\'eric} \sur{Saubion}}\email{frederic.saubion@univ-angers.fr}

 \author{\fnm{Chaïmaâ} \sur{Touhami}\footnote{Authors' names are listed in alphabetical order.}}\email{ctouhami@etud.univ-angers.fr}

\affil{\orgdiv{LERIA}, \orgname{Université d’Angers}, \orgaddress{\street{2 Boulevard Lavoisier}, \city{Angers}, \postcode{49045}, \country{France}}}

\abstract{
This paper explores a novel approach aimed at overcoming existing challenges in the realm of local search algorithms. Our aim is to improve the decision process that takes place within a local search algorithm so as to make the best possible transitions in the neighborhood at each iteration. To improve this process, we propose to use a neural network that has the same input information as conventional local search algorithms. In this paper, which is an extension of the work presented at EvoCOP2024, we investigate different ways of representing this information so as to make the algorithm as efficient as possible but also robust to monotonic transformations of the problem objective function. To assess the efficiency of this approach, we develop an experimental setup centered around NK landscape problems, offering the flexibility to adjust problem size and ruggedness. This approach offers a promising avenue for the emergence of new local search algorithms and the improvement of their problem-solving capabilities for black-box problems.  The last version of this article is published in the journal SN Computer Science (Springer).
}



\maketitle

\section{Introduction}

Local search (LS) algorithms are commonly used to heuristically solve discrete optimization problems  \cite{SLSbook}. LS algorithms are usually composed of several components: a search space, a neighborhood relation, an evaluation function, and a selection strategy. The optimization problem instance to be solved can be fully defined by its set of feasible solutions ---the decision space--- and an objective function that must be optimized. A classic and direct use of local search, when applicable, is to consider the decision space as the search space, the objective function as the evaluation function, and a natural neighborhood relation, defined from an elementary transformation function (move) such as bitflip for pseudo-Boolean problems, or induced by specific operators for permutation problems~\cite{schiavinotto2007}. 

Starting from an initial random solution, various components collaborate to drive the search towards optimal solutions. The effectiveness of this search process depends on the complexity of the problem, including factors such as deception and other structural characteristics \cite{basseur2015,jones1995,vuculescu2020,whitley2015}. The strategy to advance in the search involves selecting neighboring solutions based on their evaluations, using a wide variety of criteria. These criteria can range from simple ones, such as choosing neighbors with better or the best evaluations, to more intricate approaches involving stochastic methods. The strategy is often derived from metaheuristic frameworks based on local search such as tabu search \cite{glover1989} or iterated local search \cite{lourenco2003}. The goal is to effectively leverage the available local and partial knowledge of the landscape to identify the most promising search paths that lead to optimal solutions. 

Fitness landscape analysis \cite{malan2021} provides the optimization and evolutionary computation community with measurable characteristics (e.g, ruggedness or presence of local optima networks)  and practical tools (e.g., visualization frameworks) to examine search landscapes. It allows for the assessment of problem characteristics and the evaluation of the performance of search algorithms. In the context of combinatorial fitness landscapes, these are represented as graphs defined by a discrete search space and a neighborhood relation. A fundamental challenge lies in developing a search algorithm capable of navigating a fitness landscape to reach the highest possible fitness value. 

In general, achieving optimal solutions through a straightforward adaptive approach is quite challenging. This difficulty arises from the complex interplay among different parts of the solution, which can lead to locally optimal solutions. These local optima cannot be escaped by intensification or exploitative move strategies. Consequently, the optimization algorithm becomes stuck in a suboptimal state. The primary concern in such optimization processes is to strike a balance between exploiting promising search areas through greedy search methods and diversifying search trajectories by temporarily exploring less promising solutions.

To overcome this challenge, researchers have developed many metaheuristics \cite{sorensen2013} and even hyperheuristic schemes \cite{ozcan2008} to mix different strategies. These approaches typically incorporate parameters that allow for precise adjustment of the trade-off between exploration and exploitation. Still, they might not perform efficiently in a black-box context, as many heuristics leverage the unique properties of the given problem to solve it efficiently. Machine learning techniques have been widely used to improve combinatorial optimization solving \cite{Talbi21a} and to address the optimal configuration of solving algorithms. An approach to algorithm design known as "programming by optimization" (PbO) was introduced by Hoos \cite{Hoos12}. This paradigm encourages algorithm developers to adopt and leverage extensive design possibilities that encompass a wide range of algorithmic techniques, to optimize performance for specific categories or groups of problem instances. 

In various works, different machine learning approaches have been used either to consider offline adjustment, selection of parameters, or online control of the search process using reinforcement learning (RL) techniques (see \cite{MamaghanMMKT22} for a recent survey). The use of neural networks (NNs) in solving combinatorial optimization problems has been studied for decades \cite{TrafalisK09}, starting with the early work of Hopfield \cite{Hopfield82}. Recent applications of Graph Neural Networks (GNN) in the context of combinatorial optimization have been proposed to reach optimal solutions or to assist the solving algorithm in proving the optimality of a given solution (see \cite{CappartCK00V23} for a recent survey). 

In the traveling salesman problem, a GNN  can be used to predict the regret\footnote{In machine learning, regret measures the difference between the actual cumulative reward obtained by a learning process and the cumulative reward of an optimal strategy. It quantifies how much worse the algorithm performs compared to the best possible choice. Hence, lower regret indicates better performance.} associated with adding each edge to the solution to improve the computation of the fitness function of the LS algorithm \cite{HudsonLMP22}. Note that NNs are classically used in surrogate model-based optimization \cite{WillmesBJS03}. In \cite{SantanaLV23}, the authors introduce a GNN into a hybrid genetic search process to solve the vehicle routing problem. The GNN is used to predict the efficiency of search operators and to select them optimally in the solving process. A deep Q-learning approach has recently been proposed to manage the different stages of an LS-based metaheuristic to solve routing and job-shop problems \cite{FalknerTBS22}. High-level solving policies can often be managed by reinforcement learning in LS processes \cite{VeerapenHS13}. In this paper, our purpose is different, as we focus on building simple search heuristics for black-box problems, rather than scheduling specific operators or parameters. In particular, we assume that the learning process cannot be based on the immediate rewards that are used in RL. This motivates our choice of neuro-evolution. Note that this type of approach has recently been used to discover new genetic and evolutionary algorithms to solve continuous black-box optimization problems  \cite{lange2023discoveringGen,lange2023discoveringEvo}. In this work, we propose to use it to discover new local search algorithms for combinatorial optimization problems.

{\bf Objective of the paper} 
This study explores the potential for emerging search strategies to overcome existing challenges. 
The objective is to change how information is leveraged while retaining simple and generic search components. Considering a basic hill climber algorithm to achieve a baseline search process for solving black-box binary problem instances, our aim is to benefit from machine learning techniques to get new local search heuristics that will be built from basic search information instead of choosing a priori a predefined move heuristic (e.g., always select the best neighbor). Hence, our goal is to provide a NN with the same information as a basic LS and, after training, to use the NN as the basic move component of a simple LS process. In this work, we study different ways of representing this information, from the most basic possible representation to a more complex one that exploits an ordering relation in the neighborhood. To evaluate the efficiency of our approach, we define an experimental setup based on NK landscape problems \cite{kauffman1989}, which allows us to describe a fitness landscape whose problem size and ruggedness (determining the number of local minima) can be adjusted as parameters.  Finally, to check the robustness and generalizability of our learned strategies, we propose to evaluate them on new instances of another type of pseudo-Boolean problem, namely the quadratic unconstrained binary optimization (QUBO) problem.

\section{General framework}

\subsection{Pseudo-Boolean Optimization Problems and Local Search}
\label{sec:binopt}

Let us consider a finite set $\mathcal{X} \subseteq \{0,1 \}^N$ of solutions to a specific problem instance. These solutions are tuples of values that must satisfy certain constraints, which may or may not be explicitly provided. We evaluate the quality of these solutions using a pseudo-Boolean objective function $f_{\rm obj}: \mathcal{X} \rightarrow \mathbb{R}$. Therefore, a problem instance can be fully characterized by the pair $(\mathcal{X},f_{\rm obj})$. In terms of solving this problem, $\mathcal{X}$ is referred to as the search space.

When solving an optimization problem instance with an LS algorithm, the objective is to identify a solution $x \in \mathcal{X}$ that optimizes the value of $f_{\rm obj}(x)$. Since we are primarily concerned with maximization problems, let us note that any minimization problem can be reformulated as a maximization problem.
In this context, an optimal solution, denoted as $x^* \in \mathcal{X}$, must satisfy the condition that for every $x \in \mathcal{X}$, $f_{\rm obj}(x) \leq f_{\rm obj}(x^*)$. While exhaustive search methods, or branch and bound algorithms, guarantee the computation of an optimal solution, this is not the case with LS algorithms. However, computing optimal solutions within a reasonable time is often infeasible, leading to the use of local search algorithms within a limited budget of evaluations of $f_{\rm obj}$ to approximate near-optimal solutions.

LS algorithms operate within a structured search space, thanks to a fixed-sized neighborhood function denoted as $\mathcal{N}:\mathcal{X}\rightarrow 2^\mathcal{X}$. This function assigns a set of neighboring solutions $\mathcal{N}(x) \subseteq \mathcal{X}$ for each solution $x\in\mathcal{X}$. To maintain a fundamentally generic approach to LS, we assume that $\mathcal{N}$ is defined using basic flip functions, $flip_i: \mathcal{X} \to \mathcal{X}$, where $i \in \llbracket 1, N \rrbracket$, and $flip_i(x)$ is equal to $x$ except for the $i^{th}$ element, which is changed from $0$ to $1$ or vice versa. In this case, $\mathcal{N}(x) = \{ flip_i(x) \mid  i \in \llbracket 1,N \rrbracket \}$. Starting from an initial solution, often selected randomly and denoted as $x_0$, LS constructs a path through the search space based on neighborhood relationships. This path is typically represented as a sequence of solutions bounded by a limit (horizon) $H$, denoted as $(x_0,x_1,\dots,x_H)$, where for each $i\in \llbracket 0, H-1 \rrbracket $, $x_{i+1}\in \mathcal{N}(x_i)$. Let us denote ${\cal P} = \mathcal{X}^*$ the set of all paths (i.e., the set of all possible sequences built on ${\cal X}$). 

In addition to the neighborhood function, this sequence of solutions is determined by a strategy, often involving the use of a fitness function $f$. For example, hill climbing algorithms select the next solution on the path based on a simple criterion: $\forall t \in \llbracket 0, H-1\rrbracket, f(x_t) < f(x_{t+1})$, where $f(x_t)$ represents the fitness evaluation of solution $x_t$. The process of choosing the next solution on the search path is referred to as a move. We denote $x_{t+1} = x_t \oplus flip_i$ the move  that corresponds to $x_{t+1}=flip_i(x_t)$. 





\subsection{Local Search as an Episodic Task Process \label{sec:episodicLS}}

According to previous remarks, an LS process can be modeled by a sequence of actions performed on states. Let us describe all the components of this episodic task process in the following subsections.

\subsubsection{States}

In this paper, we consider only LS processes that do not involve memory that records past decisions. In this work, we assume that a state $s \in \mathcal{X} \times 2^\mathcal{X}$ is fully described by a current solution $x \in {\cal X}$ and its neighborhood $\mathcal{N}(x)$. $\mathcal{S}$ corresponds to the set of states that can be reached during the search. We are in the context of episodic (discrete state) tasks with a terminal state (the end of the search, e.g. fixed by a maximal number of moves $H$).

\subsubsection{Observations \label{sec:observations}}

We introduce the notion of observation of a state as a function corresponding to an abstraction of the real search states in order to gather only useful information for the considered local search strategy.

In this work, we propose to study different observation functions that will be used by local search strategies to make decisions. Given $s \in \mathcal{S}$, with $s = (x, \mathcal{N}(x))$, with $x$ the current solution encountered by the local search and  $\mathcal{N}(x) = \{ flip_i(x) \mid  i \in \llbracket 1,N \rrbracket \}$ its  one-flip neighborhood, we define four observation functions from $o^k : \mathcal{S}$ to $\mathbb{R}^{N \times d }$ using  superscript $k$ to distinguish between them. The notation $o^k(x)_i$ is used to denote the $i^{th}$ line of $o^k(x)$ (in case of $d=1$ this line if of course the $i^{th}$ value of the corresponding 1D vector).

\begin{enumerate}

\item The first observation function abstracts the state as the variation of the fitness function for each possible flip of the values of $x$ ($d$ is set to 1). 

\begin{equation*}
o^1(x) = 
\begin{pmatrix}
\Delta_1(x) \\
\Delta_2(x) \\
\vdots     \\
\Delta_N(x) \\
\end{pmatrix},
\end{equation*}

\noindent where $\Delta_i(x) = f_{{\rm obj}}(x) - f_{{\rm obj}}(flip_i(x))$.

\item The second observation function describes the state with a matrix with $N$ rows and $d=2$ columns. It gives the information of the value of the fitness function for the current solution as well as the value of the fitness function for each possible solution in $\mathcal{N}(x)$.

\begin{equation*}
o^2(x) = 
\begin{pmatrix}
f_{{\rm obj}}(x) & f_{{\rm obj}}(flip_1(x)) \\
f_{{\rm obj}}(x) & f_{{\rm obj}}(flip_2(x)) \\
\vdots  & \vdots   \\
f_{{\rm obj}}(x) & f_{{\rm obj}}(flip_N(x)) \\
\end{pmatrix},
\end{equation*}

Let us note that the same value $f_{{\rm obj}}(x)$ is duplicated in each line of $o^2(x)$ since the NN (see Section \ref{sec:NeuroLS}) is applied to each line $o(x)_i$.

\item The third observation function $o^3 : \mathcal{S} \rightarrow \mathbb{R}^{{N}\times{d}} $, with $d=1$ is based on a ranking of the solutions associated with positive, negative or null variations of the fitness function for each possible flip of the values of $x$. We define $N^+(x)$ and $N^-(x)$   the numbers of solutions in $\mathcal{N}(x)$ with respectively strictly greater and lower fitness than $x$.

Hence, $o^3(x)$ is  vector of size $N$, such that 
for $i = 1, \dots, N$:
\begin{itemize}
\item $o^3(x)_i = rank^+(\Delta_i(x)) / N^+(x)$,  if $\Delta_i(x) > 0$;
\item $o^3(x)_i = rank^-(\Delta_i(x))/ N^-(x) $,  if $\Delta_i(x) < 0$;
\item $o^3(x)_i = 0$, if $\Delta_i(x) = 0$.
\end{itemize}

$rank^+$ is a function that associates to a strictly positive value $\Delta_i(x)$ its rank among all strictly positive $\Delta_i(x)$ values, assigning a value from $\{1, \dots, N^+(x) \}$. The smallest positive $\Delta_i(x)$ receives the value of 1, while the largest positive $\Delta_i(x)$ receives the value $N^+(x)$ (with ties broken randomly).

Symmetrically, $rank^-$ is a function that associates to a strictly negative value $\Delta_i(x)$ its rank among all strictly negative $\Delta_i(x)$ values, assigning a value from $\{-N^-(x), \dots, -1 \}$. The smallest negative $\Delta_i(x)$ receives the value $-N^-(x)$, while the largest negative $\Delta_i(x)$ receives the value $-1$.

\textit{Example 2}
Let us consider the following situation with $x \in S$ and $N=6$ : $\Delta_1(x)=1,\Delta_2(x)=4,\Delta_3(x)=-2,\Delta_4(x)=-5,\Delta_5(x)=0,\Delta_6(x)=-7$. 
We have thus $rank^+(\Delta_1(x))=1,rank^+(\Delta_2(x))=2,rank^-(\Delta_3(x))=-1,rank^-(\Delta_4(x))=-2,rank^-(\Delta_6(x))=-3$. Hence, the corresponding observations are $o^3(\Delta_1(x))=\frac{1}{2},o^3(\Delta_2(x))=1,o^3(\Delta_3(x))=-\frac{1}{3},o^3(\Delta_4(x))=-\frac{2}{3},o^3(\Delta_5(x))=0,o^3(\Delta_6(x))=-1$.

This rank transformation always returns values in the range $[-1,1]$ and keeps the information that a flip move can deteriorate or improve the current fitness solution. An interesting property of this transformation, is that it is independent of a change in the scale of the objective function. $o^3(x)$ remains unchanged even if $f_{{\rm obj}}$ is multiplied by a strictly positive real value $\lambda$. 

\item The fourth observation function $o^4 : \mathcal{S} \rightarrow \mathbb{R}^{N \times d }$, with $d=2$ uses the same ranking transformation as $o^3$, in addition to a z-score transformation of the $\Delta_i(x)$ values. 
We define the z-score of $\Delta_i(x)$ as $Z(\Delta_i(x)) = \frac{\Delta_i(x) - \mu(x)}{\sigma(x)}$, with $\mu(x)$ and $\sigma(x)$ are the mean and standard deviation, respectively, of the set of $\Delta_i(x)$ values for $i = 1, \dots, N$.  This z-score indicates how many standard deviations a given fitness variation is above or below the mean of possible fitness variations. Thus, 

\begin{equation*}
o^4(x) = 
\begin{pmatrix}
o^3(x)_1 & Z(\Delta_1(x))\\
o^3(x)_2 & Z(\Delta_2(x))\\
\vdots  & \vdots   \\
o^3(x)_N & Z(\Delta_N(x))\\
\end{pmatrix}.
\end{equation*}
\end{enumerate}


\subsubsection{Deterministic policy and actions}

 Following a reinforcement learning-based description, a local search process can be encoded by  a policy $\pi : \mathbb{R}^{N\times d} \to {\cal A}$ where ${\cal A}$ is a set of actions. Here we consider deterministic policies, i.e. for each $o(x) \in \mathbb{R}^{N\times d}$, there exists one and only one action $a \in {\cal A}$, such that $\pi(o(x)) = a$. Note that the policy can be parameterized by a parameter vector $\theta$. The set of all policies is  $\Pi = \{\pi_\theta \mid \theta \in \Theta \}$ where $\Theta$ is the parameter space. 

In our context, we consider only actions that are bitflips $flip_i$ defined in Section~\ref{sec:binopt}. Hence, $\mathcal{A} = \{ flip_i \mid i \in \llbracket 1, N \rrbracket \}$. Note that of course, other actions can be introduced to fit specific strategies. 

\subsubsection{Transitions and trajectory}

We consider a transition function $\delta : {\cal X} \times {\cal A} \to {\cal X}$ such that $\delta(x,a)= x \oplus a$. This function specifies the change in state of the environment in response to the chosen action. Note that the transition is used to update the current state and compute the next state of the LS process. An LS run can be fully characterized by the search trajectory that has been produced from an initial starting solution. 

Given an instance $I=({\cal X},f_{{\rm obj}})$, an initial solution $x_0\in {\cal X}$, a policy $\pi_\theta$, and a horizon $H$, a trajectory $T(x_0,\pi_\theta,H,I)$ is a sequence $(x_0,a_0,x_1,a_1,\dots,x_{H-1},$ $a_{H-1},x_H)$ such that $(x_0,\dots,x_H) \in {\cal P}$ is an LS path (the multiset $\{x_0,\dots,x_H\}$ will be denoted $P(T(x_0,\pi_\theta,H,I))$), $\forall i \in \llbracket 0,H-1 \rrbracket, a_i=\pi_{\theta}(o(x_i))$ and  $\forall i \in \llbracket 1,H \rrbracket, x_i=\delta(x_{i-1},a_{i-1})$. Note that the trajectory is defined with regard to solutions belonging to ${\cal X}$, while the policy operates on the space of observations obtained from these solutions. 

\textit{Example 1.} In order to illustrate our framework, let us consider a basic hill climber (HC) that uses a simple best-improve strategy using objective function $f_{{\rm obj}}$ as fitness function. In the following, we only consider LS processes that do not involve memory that records past decisions. Hence, a state of a HC is fully described by a current solution $x \in {\cal X}$ and its neighborhood $\mathcal{N}(x)$. An observation will abstract the state as the variation of the fitness function for each possible flip of the values of $x$, $o^2(x) = (\Delta_1(x),\dots, \Delta_N(x)) \in \mathbb{R}^N$. In a best-improve HC process, the search stops when a local optimum is reached and no improving move can be performed. We consider ${\cal A} =\{ flip_i \mid i \in \llbracket 1, N\rrbracket \}$. Then, we define the policy $\pi_{HC}(o^1(x)) = \argmax_{a \in {\cal A }} (f_{{\rm obj}}(\delta(x,a))-f_{{\rm obj}}(x))$. Note that when a local optimum is reached, the search gets stuck.

\subsubsection{Reward}

\color{black}

Compared to classic reinforcement schemes, where a reward can be assigned after each action, in our context, the reward will be computed globally for a given trajectory. Note that if we considered only the best-improvement strategy, then the reward could be assigned after each move to assess that the best move has been selected. Unfortunately, such a strategy will only be suitable for simple smooth unimodal problems. Hence we translate this reward as a function $R(x_0,\pi_\theta,H,I)=\max_{x \in P(T(x_0,\pi_\theta,H,I))} f_{{\rm obj}}(x)$.

Our objective is to maximize the maximum score encountered by the agent during its trajectory, and not the sum of local fitness improvements obtained during its trajectory. We are therefore not in the case of learning a Markov decision process. This is why classical reinforcement learning algorithms such as Q-learning or policy gradient are not applicable in this context.

This justifies our choice of neuro-evolution where the policy parameters will be abstracted by a neural network  that will be used to select the suitable action. The parameters of this neural network will be searched by means of an evolutionary algorithm according to a learning process defined below.

\subsection{Policy Learning for a Set of Instances \label{sec:NKgenerator}}

In this paper, we focus on NK landscapes as pseudo-Boolean optimization problems. The NK landscape model was introduced to describe binary fitness landscapes \cite{kauffman1989}. The characteristics of these landscapes are determined by two key parameters: $N$, which represents the dimension (number of variables), and $K$ (where $K < N$), which indicates the average number of dependencies per variable and, in turn, influences the ruggedness of the fitness landscape. An NK problem instance is an optimization problem represented by an NK function. We use random NK functions to create optimization problem instances with adjustable search landscape characteristics. This adjustment will be achieved by varying the parameters $(N, K)$, thus generating diverse search landscapes. Hence, we consider ${\cal NK}(N,K)$ as a distribution of instances generated by a random NK function generator. 

In order to achieve our policy learning process, we must assess the performance of a policy as $F(\pi_\theta,{\cal NK}(N,K),H)= \mathbb{E}_{ I \sim {\cal NK}(N,K), x_0 \sim {\cal X}} [R(x_0,\pi_\theta,H,I)] $ where  $x_0 \sim {\cal X}$ stands for a uniform selection of an element in ${\cal X}$. However, since this expectancy cannot be practically evaluated, we rely on an empirical estimator computed as an average of the scores obtained by the policy $\pi_{\theta}$ for a finite number $q$ of instances $I_1, \dots, I_q$ sampled from the distribution ${\cal NK}(N,K)$ and a finite number $r$ of  restarts $x^{(1)}_0, \dots, x^{(r)}_0$ drawn uniformly in ${\cal X}$:

\begin{equation}
\label{eq:empiricalScore}
    \bar{F}(\pi_\theta,{\cal NK}(N,K),H)= \frac{1}{q r  } \sum_{ i = 1}^q \sum_{ j = 1}^r R(x^{(j)}_0,\pi_\theta,H,I_i).
\end{equation}

\begin{figure}[h]
    \centering

        \includegraphics[width=0.8\textwidth]{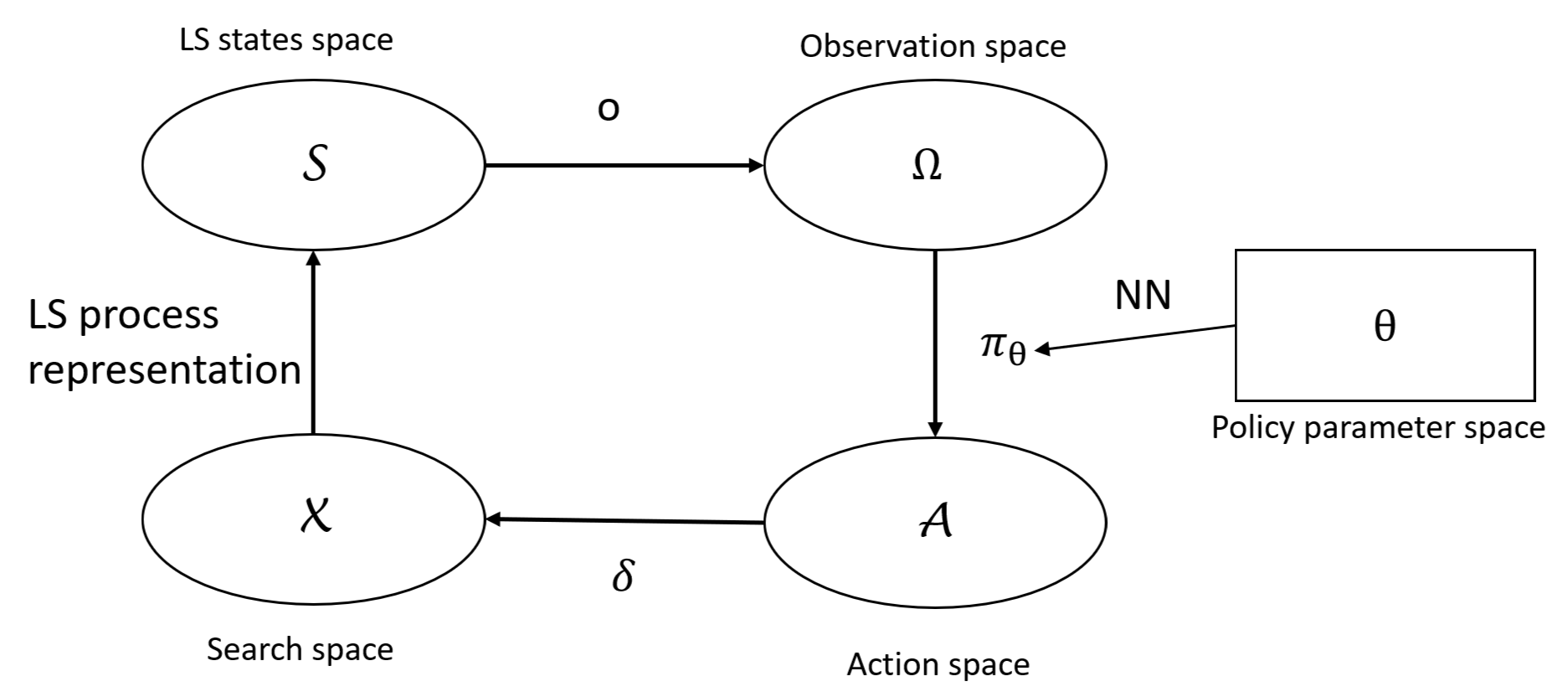}
    \caption{Global View of the Process. \label{fig:global}}
\end{figure}

Figure \ref{fig:global} highlights our general learning methodology and the connections between the LS process at the instance solving level and the policy learning task that will be achieved by a neural network presented in the next section.   

\section{Deterministic local search policies for pseudo-Boolean optimization}
\label{sec:NeuroLS}

In this paper, our objective is to compare an LS policy learned by neuro-evolution, with three different baseline LS algorithms. To ensure a fair comparison among all the algorithms, all the different strategies take as input the same type of information, i.e. the fitness of the current solution $x$ and the fitness of the solutions in its one-flip neighborhood $\mathcal{N}(x)$. All of these strategies are deterministic and memoryless.  The set of possible actions available to the different strategies always remains ${\cal A} =\{ flip_i \mid i \in \llbracket 1, N\rrbracket \}$.

\subsection{Neural network local search policy \label{sec:nn}}

We introduce a deterministic LS policy $\pi_{\theta} : \mathbb{R}^{N\times d} \rightarrow \mathcal{A}$, called Neuro-LS, which uses a neural network $g_{\theta}$, parametrized by a vector of real numbers $\theta$.  
When in a state $x$, this neural network takes as input an observation matrices $o(x) \in \mathbb{R}^{N \times d}$  and gives as output a vector $g_{\theta}(o(x))$ of size $N$ whose component $g_{\theta}(o(x))_i$ corresponds to a preference score associated to each observation $o_i$. Then, the action $a_i$ corresponding to the highest score $g_{\theta}(o)_i$ is selected. 

In the experimental section of this paper, we will compare the impact on the results of using the different types of observation matrices as input for the neural network (see Section \ref{sec:observations}).

A desirable property of this neural network policy is to be \textit{permutation equivariant} with respect to the input vector of observations, which is a property generally entailed by a local search algorithm, in order to make its behavior consistent for solving any type of instance. Formally, an LS algorithm is said \textit{to be invariant to permutations} in the observations if for any permutation $\sigma$ on $\llbracket1, N\rrbracket$, we have $a_{\sigma(i)}= \pi_{\theta}(o(x)_{\sigma(1)}, o(x)_{\sigma(2)}, \dots, o(x)_{\sigma(N)})$. 
As an example, the basic hill climber HC defined with Example 1 in Section \ref{sec:episodicLS} has this property.

In order to obtain this property for Neuro-LS, the same regression neural network  $g_{\theta} : \mathbb{R}^d \rightarrow 1 $  is applied to each row $o(x)_i$ of the observation matrix $o(x) \in \mathbb{R}^{N \times d}$. Note that this is a simplified version of  architecture presented at EvoCOP2024 in \cite{goudet2024emergence}. In this version, each variable is processed independently, which make the interpretation of the function learned by the neural network more explainable  (see Section  \ref{sec:emerging}).

Formally, given $o(x)_i$ as input, the output of the neural network is calculated as  

\begin{equation}
g_{\theta}(o(x)_i) = L_{H+1} \circ \sigma \circ L_{H} \circ \dots \circ \sigma \circ L_{1}(o(x)_i),
\label{eq:fcm_neural}
\end{equation}

\noindent where $L_{h} : \mathbb{R}^{n_{h-1}} \rightarrow \mathbb{R}^{n_h}$ is an affine linear map defined by $L_{h}({x}) = {W}_{h} \cdot {x} + {b}_{h}$ for given $n_h \times n_{h-1}$ dimensional weight matrix ${W}_{h}$ (with coefficients $\{w^{h}_{k,l}\}_{1 \leq k \leq n_h \atop 1 \leq l \leq n_{h-1}}$), $n_h$ dimensional bias vector ${b}_{h}$ (with coefficients $\{b^{h}_{k}\}_{1 \leq k \leq n_h}$) and $\sigma : \mathbb{R}^{n_h} \rightarrow ]-1,1[^{n_h}$\footnote{$]-1,1[$ is used to denote the open interval included in $\mathbb{R}$, while  $\llbracket 1, N\rrbracket$ in an integer values interval.}  the element-wise nonlinear activation map defined by $\sigma({z}) := (tanh(z_1),...,tanh(z_{n_h}))^{\intercal}$. We denote by $\theta$, the set of all weight matrices and bias vector of  the neural network $g_{\theta}$ : $\theta := \{ ({W}_{1}, {b}_1), ({W}_{2}, {b}_2), \dots, ({W}_{{H+1}}, {b}_{H+1})\}$.

\begin{figure}[h]
    \centering
    \includegraphics[width=0.8\textwidth]{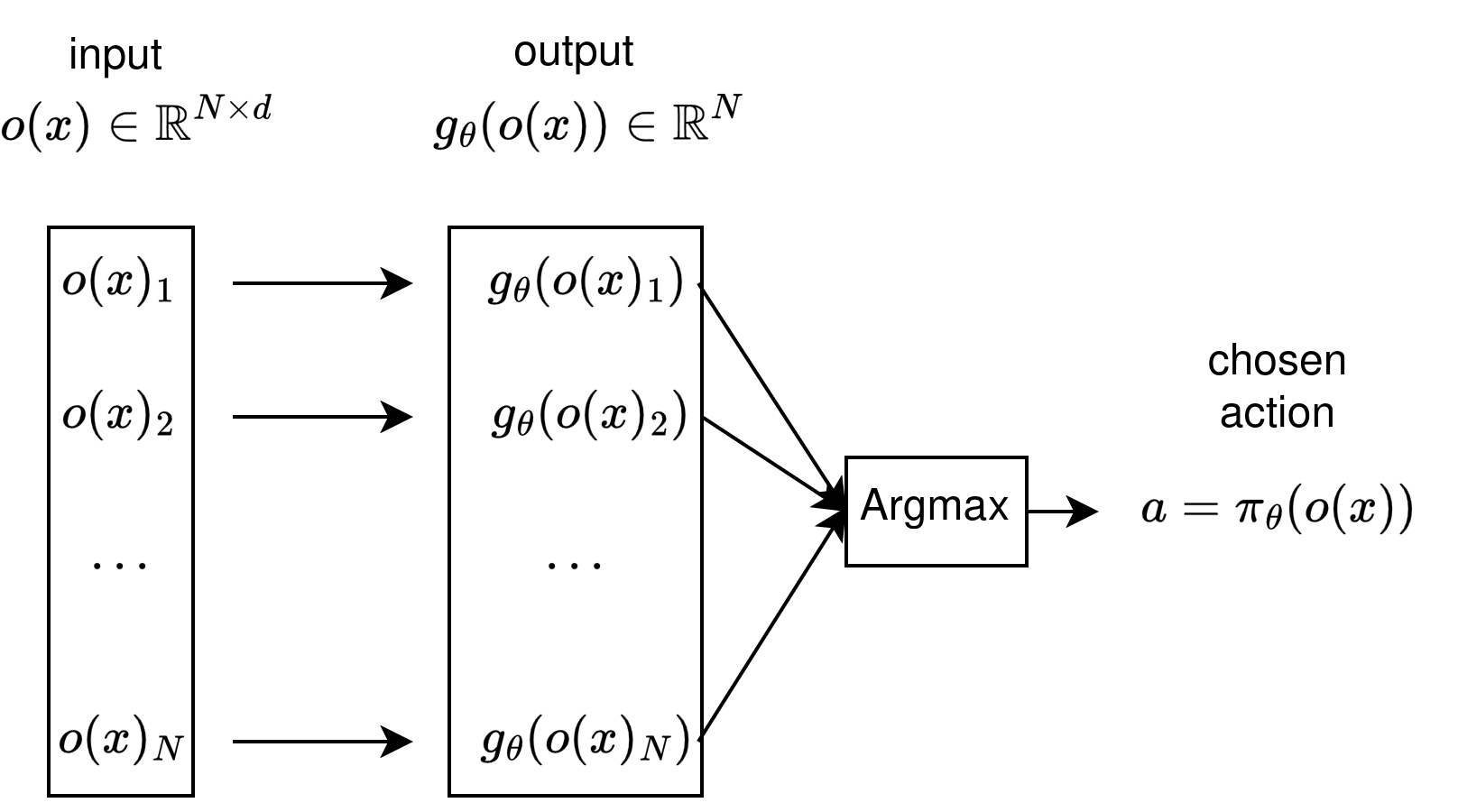}
    \caption{ Neuro-LS policy \label{fig:neural_network_policy}.}
\end{figure}

The Neuro-LS policy $\pi_{\theta} : \mathbb{R}^{N \times d} \rightarrow \mathcal{A}$ using the neural network $g_{\theta}$ is displayed in Figure 
\ref{fig:neural_network_policy}.

Given a vector of observation $o(x) \in \mathbb{R}^{N \times d}$, the neural network outputs a vector $g_{\theta}(o(x)) = (g_{\theta}(o(x)_ 1), \dots, g_{\theta}(o(x)_N)) \in \mathbb{R}^{N}$. Then, the action $a = \argmax_{i\in \llbracket 1,N \rrbracket} g_{\theta}(o(x)_i)$ is returned by $\pi_{\theta}$.

 Note that the number of parameters in  $g_{\theta}$ does not depend on the size $N$ of the observation vector, allowing the same Neuro-LS strategy to be used for pseudo-Boolean optimization problems of different sizes (see Section \ref{sec:outOfDistrib} below).

 \subsection*{Neuro-evolution with CMA-ES}
\vspace{-0.5 em}

The neural network policy $\pi_{\theta}$ is characterized by a set of parameters denoted as $\theta$. The optimization goal is to maximize the estimated score $\bar{F}(\pi_\theta, {\cal NK}(N,K),H)$. This poses a stochastic black-box optimization problem within the real-valued search space $\mathbb{R}^{\mid \theta \mid }$.
To tackle this problem, we propose to use the covariance matrix adaptation evolution strategy (CMA-ES)  \cite{hansen2001completely}, which stands out as one of the most powerful evolutionary algorithms for addressing such black-box optimization problems \cite{muller2018challenges}.

The principle of CMA-ES is to iteratively test new generations of real-valued parameter vectors $\theta$ (individuals). Each new generation of parameter vectors is stochastically sampled according to a multivariate normal distribution. The mean and covariance matrix of this distribution are incrementally updated, so as to maximize the likelihood of previously successful candidate solutions. Thanks to the use of a stepwise-adapted covariance matrix, the algorithm is able to quickly detect correlations between parameters, which is an important advantage when optimizing the (many) parameters of the neural network policy. Another advantage of CMA-ES is that it relies on a ranking mechanism of the estimated scores $\bar{F}$ given by the Equation \eqref{eq:empiricalScore} for the different individuals of the population, rather than on their absolute values, making the algorithm more robust to stochastic noise related to the incertitude on the estimation of the performance score $F$ with a finite number of trajectories.

\subsection{Basic local search policies \label{sec:basicLS}}

We compare the neural network policy above with three \textit{basic} local search policies which have been extensively studied in the literature \cite{ochoa2010first,basseur2013hill}. All these strategies take as input the same vector of observation as the neural network policy and return an action $a \in \mathcal{A}$. These three policies are two hill climbers as well as a $(1,\lambda)$-evolution strategy \cite{beyer2001theory} used as a local search \cite{tari2021use}.  They are made deterministic using a pseudo-random number generator $h$ whose seed is determined with a hash function from the current state $x$ encountered by the LS.  

\subsubsection{Best improvement hill climber {\small [+jump]} (${\rm BHC}^{+}$).}  
 
 This strategy  always selects the action $a_i=flip_i \in \mathcal{A}$ in such a way that $f(a_i(x))-f(x)$ is maximized, provided there is at least one action $a_i$ that strictly improves the score. If $\forall i,\ f(a_i(x))-f(x) \leq 0$, then this strategy performs a random jump by choosing a random action $a \in \mathcal{A}$ using the pseudo-random number generator $h(x)$.


 \subsubsection{First improvement hill climber {\small [+jump]}  (${\rm FHC}^{+}$).}  This strategy iterates through all actions in $\mathcal{A}$ in random order and selects the first  action $a_i$ leading to a strictly positive score improvement, i.e. such that $f(a_i(x))-f(x) > 0$. Similar to the ${\rm BHC}^{+}$ strategy, if $\forall i,\ f(a_i(x))-f(x) \leq 0$, it performs a random jump.

 \subsubsection{$(1,\lambda)$-evolution strategy ($(1,\lambda)$-ES).}

This strategy randomly evaluates $\lambda$ actions in  ${\cal A}$ and chooses the one that yields the best score, even if it results in a deteriorating move. $\lambda$ is a method hyperparameter that will be calibrated to maximize the estimated score $\bar{F}$ for each type of NK landscape instance as detailed in the next section.


\section{Computational experiments} \label{sec:experimentation}


In this section, we target four aspects of Neuro-LS experimentally: 
\begin{enumerate}
    \item impact of the different observation functions used as inputs to the neural network on the performance of the Neuro-LS strategy,
    \item performance of Neuro-LS compared to the baseline LS strategies presented in the last subsection for NK landscape problems of different sizes and ruggedness,
    \item robustness of the learned Neuro-LS strategy for other types of instances coming from a different pseudo-Boolean problem (QUBO),
    \item study of the emergent strategies discovered by Neuro-LS at the end of its evolutionary process. 
\end{enumerate}

First, we discuss the experimental setting. Then, we follow the classical steps of a machine learning workflow: subsection \ref{sec:training} describes the Neuro-LS training process; it will allow us to select different emerging strategies for each type of NK landscape on validation sets, which we will then compare with the different baseline LS strategies on test sets  coming from the same distribution of instances (subsection \ref{sec:test}) or from an other distribution of instances (subsection \ref{sec:outOfDistrib}).
Finally, an in-depth analysis of the best emerging strategies discovered by Neuro-LS will be performed in  subsection \ref{sec:emerging}.

\subsection{Experimental settings}

In these experiments, we consider independent instances of NK-landscape problems. Twelve different scenarios with $N \in \{32, 64, 128\}$ and $K \in \{1, 2, 4, 8\}$ are considered. For each scenario, three different sets  of  instances are sampled independently from the   ${\cal NK}(N,K)$ distribution described in Section \ref{sec:NKgenerator}: a training set, a validation set and a test set.

For the resolution of each instance, given a random starting point $x_0 \in \mathcal{X}$, each LS algorithm performs a trajectory of size $H = 2 \times N$ (iterations) and returns the best solution found during this trajectory.  The experiments were performed on a computer equipped with a 12th generation Intel® CoreTM i7-1265U processor and 14.8 GB of RAM.  

Neuro-LS is implemented in Python 3.7 with Pytorch 1.4 library.\footnote{The program source code,  and benchmark instances  are available at the url \url{https://github.com/Salim-AMRI/NK_Landscape_Project.git}.} For all experiments with different values $N$ and $K$, we use the same architecture of the neural network composed of two hidden layers of size 10 and 5, with a total of $\mid \theta \mid= 81$ parameters to calibrate when $d=1$ (when using observation functions $o^1$ and $o^3$) and $\mid \theta \mid= 91$ parameters to calibrate when $d=2$ (when using observation functions $o^2$ and $o^4$).

To optimise the weights of the neural network, we used the  CMA-ES algorithm of the pycma library \cite{hansen2019pycma}.  The multivariate normal distribution of CMA-ES is initialized with mean parameter $\mu$ (randomly sampled according to a unit normal distribution) and initial standard deviation $\sigma_{\rm init}~=~0.2$.


\subsection{Neuro-LS training phase \label{sec:training}}


For each NK-landscape configuration and for each of the four observation functions used to compute the input of the neural network (see Section \ref{sec:observations}), we run 10 different  training processes of Neuro-LS with CMA-ES, and a maximum number of 100 generations of CMA-ES, to optimize the empirical score $\bar{F}$ defined by \eqref{eq:empiricalScore}, computed as an average of the best fitness scores obtained  for 100 trajectories ($r=10$ independent random restarts for each of the $q = 10$ training instances).

\begin{figure}[h]
\centering
\includegraphics[width=0.9\textwidth]{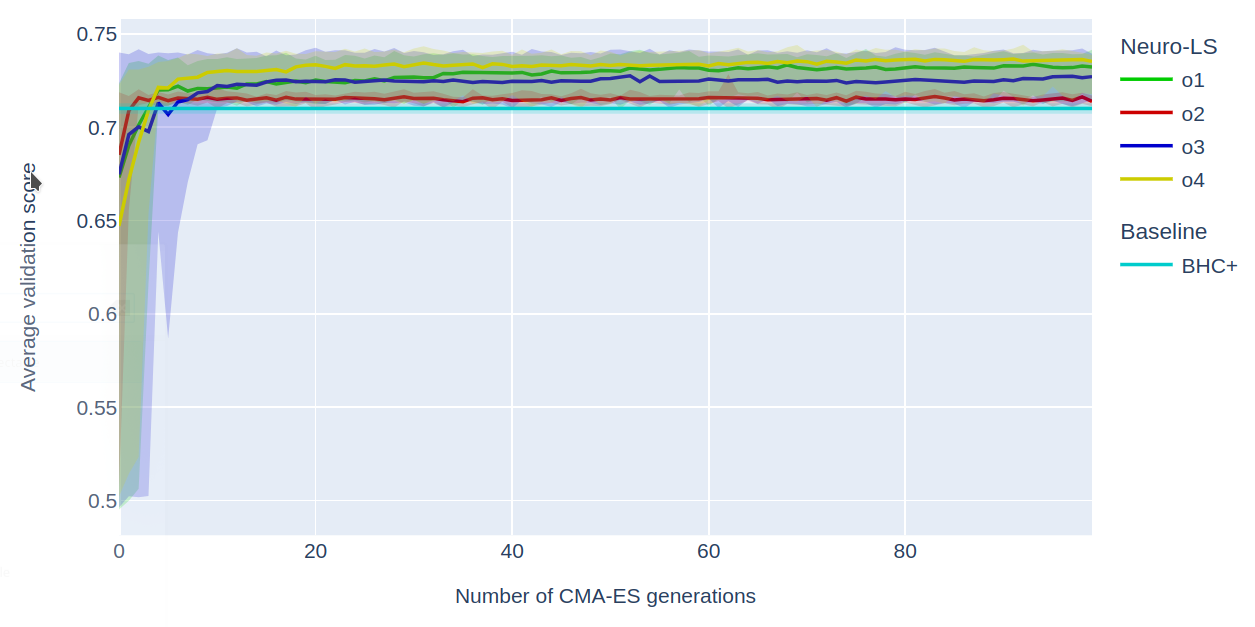}
\caption{Evolution of the average score on the validation set for NK instances with $N = 64$ and $K=8$ over the generations of CMA-ES obtained by Neuro-LS with the four different observation matrices as input.  \label{fig:training}}  
\end{figure}

Figure \ref{fig:training} displays the results of 10 independent neuro-evolution training processes performed on the NK landscape instances with $N = 64$ and $K=8$ for the different observation functions.

At each generation, the 10 training instances are regenerated to avoid over-fitting. Then, CMA-ES samples a population of 17 individuals (17 vectors of weights $\theta$  of the neural network), and  the best Neuro-LS strategy of the population on the training set is evaluated on the 10 instances of the validation set (with 10 independent restarts per instance). 

The evolution of the average score obtained on the Neuro-LS validation sets, considering as input the observation matrices $o^1$, $o^2$, $o^3$ and $o^4$  are respectively indicated with green, red, blue and yellow lines in Figure \ref{fig:training}. 

The light blue  line is a reference score. It corresponds to the average score $\bar{F}$ obtained by the ${\rm BHC}^{+}$ local search strategy (see Section \ref{sec:basicLS}) on the same validation set.

First, when comparing the validation curves of Neuro-LS and ${\rm BHC}^{+}$, we observe that all the Neuro-LS curves progress over generations, and eventually surpass ${\rm BHC}^{+}$ when evaluated on the same set of  validation instances. This finding highlights that, once trained, the Neuro-LS method is able to find solutions more efficiently compared to the baseline ${\rm BHC}^{+}$ local search for this type of instances.

The strategy, which obtains the best results on average over the validation set, is the strategy that uses the $o^4$ observation function (let us recall that $o^4$ extracts information from both the rank of improving and deteriorating neighbors and the z-scores of fitness variations for each possible flip). This strategy learns faster than the other strategies and obtains more consistent scores across the different training runs.

In Figure \ref{fig:training}, the colored area around each solid line is delimited by the minimum and maximum scores obtained by each strategy on the validation set for each generation.
This figure demonstrates significant variability in the results, highlighting the diversity in the performance of the emerging strategies. 



However, this variability does not present a challenge in our context, since only the run with the best results on the validation set for each strategy is selected for the testing phase presented in the next subsection.

\subsection{Test phase \label{sec:test}}

In this phase, we performed a series of evaluations to assess whether the best Neuro-LS strategies, selected based on the validation set for each configuration of NK landscape, continue to perform well in new test instances that are independently sampled from the same ${\cal NK}(N,K)$ distribution.

Table \ref{tab:test} summarizes the average score  obtained by the four variants of Neuro-LS, with the different observation functions $o^1$, $o^2$, $o^3$ and $o^4$, and the three other competing methods, namely the ${\rm BHC}^{+}$, ${\rm FHC}^{+}$ and (1, $\lambda$)-ES algorithms, presented in Section \ref{sec:basicLS}. 

The strategy (1, $\lambda$)-ES has a hyperparameter $\lambda$ that we calibrated in the range $\llbracket 1, N \rrbracket$ on the training instances, for each $(N,K)$ configuration.


To perform a fair comparison between the different strategies, we compute an average estimated score $\bar{F}$ on the same 100 test instances. For each instance, we use the same starting point to compute the trajectory produced by each LS strategy.\footnote{For this evaluation test, we only perform one restart per instance, to avoid any dependency between the different executions that might take place on the same instance. It allows us to obtain a distribution of 100 independently and identically distributed scores for each strategy and each NK configuration.}

In Table \ref{tab:test}, the best average result obtained among the mentioned methods for each configuration $(N,K)$ of test instances is highlighted in bold. 

Results underlined indicate significant better results obtained by a Neuro-LS strategy on average compared to all the baseline strategies (p-value below 0.001), measured with a one-tailed Student t-test without assuming equal variance and with a Bonferroni correction to correct the significance threshold for multiple comparisons.\footnote{The normality condition required for this test was first confirmed using a Shapiro statistical test on the empirical distributions of 100 iid scores obtained by each strategy.}



\begin{table}
\centering
\resizebox{0.95\textwidth}{!}{
\begin{tabular}{c c |c|c|c|c|c|c|c}
\hline
\multicolumn{2}{c|}{Instances} & \multicolumn{7}{c}{Methods} \\ 
\cline{1-9} 
\multirow{2}{*}{N} & \multirow{2}{*}{K} & \multirow{2}{*}{${\rm BHC}^{+}$} & \multirow{2}{*}{${\rm FHC}^{+}$} & \multirow{2}{*}{(1, $\lambda$)-ES} & \multicolumn{4}{c}{Neuro-LS} \\
\cline{6-9}

  &  &  &  &  & $o^1$ & $o^2$ & $o^3$ & $o^4$  \\ \hline\hline
 
32 & 1 & 0.694 & 0.688 & 0.695 & \textbf{0.699} & \textbf{0.699} & \textbf{0.699} & \textbf{0.699} \\ \hline
32 & 2 & 0.716 & 0.714 & 0.717 & \textbf{0.719} & \textbf{0.710} & \textbf{0.719} & \textbf{0.719} \\ \hline
32 & 4 & 0.717 & 0.723 & 0.713 & \underline{\textbf{0.742}} & 0.724 &  \underline{0.739} & \underline{0.736} \\ \hline
32 & 8 & 0.705 & 0.716 & 0.707 & \underline{0.730} & 0.718 &  \underline{\textbf{0.731}} & \underline{0.729} \\ \hline\hline
64 & 1 & 0.700 & 0.694 & 0.696 & \textbf{0.701} & 0.699 & \textbf{0.701} & \textbf{0.701} \\ \hline
64 & 2 & 0.716 & 0.713 & 0.712 & \textbf{0.717} & \textbf{0.717} & \textbf{0.717} & \textbf{0.717} \\ \hline
64 & 4 & 0.718 & 0.726 & 0.714 & \underline{0.743} & 0.726 &  \underline{0.742} & \underline{\textbf{0.744}} \\ \hline
64 & 8 & 0.706 & 0.714 & 0.707 & \underline{0.737} & 0.726 &  \underline{\textbf{0.739}} & \underline{0.738} \\ \hline\hline
128 & 1 & \textbf{0.699} & 0.694 & 0.696 & \textbf{0.699} & \textbf{0.699} & \textbf{0.699} & \textbf{0.699}  \\ \hline
128 & 2 & 0.715 & 0.709 & 0.710 & \textbf{0.716} & 0.713 & \textbf{0.716} & \textbf{0.716} \\ \hline
128 & 4 & 0.723 & 0.726 & 0.717 & \underline{0.738} & 0.726 &  \underline{\textbf{0.740}} & \underline{\textbf{0.740}} \\ \hline
128 & 8 & 0.710 & 0.721 & 0.705 & \underline{0.739} & 0.720 &  \underline{\textbf{0.740}} & \underline{0.743} \\ \hline
\end{tabular}
}
\caption{Average results on test instances obtained by the baseline local search strategies and the Neuro-LS strategies with different type of observations given as input for different NK landscape configurations. Best result on each line is written in bold. Results underlined indicate significant better results obtained by a Neuro-LS strategy on average compared to all the baseline strategies (p-value below 0.001), measured with a one-tailed Student t-test without assuming equal variance and with a Bonferroni correction to correct the significance threshold for multiple comparisons.  \label{tab:test}}
\end{table}


Table \ref{tab:test} shows that all the Neuro-LS variants always obtains equal or better results than the baseline algorithms for all the configurations of NK landscape, but the difference in score is only really significant when $K=4$ and $K=8$ for all values of $N$.
It means that our learned Neuro-LS strategies become more effective than other methods when the landscape  is more rugged.

This score improvement compared to the other baseline methods, obtained with the same budget of $H=2\times N$ iterations performed on each instance, can be attributed to a more efficient exploration of the search space as $K$ increases (as seen in Section \ref{sec:emerging}).

Among the Neuro-LS strategies, we observe that the strategy using the second observation function $o^2$ gets the worst scores on the validation sets.  This is because the neural network struggles to extract relevant information using the rawest possible information about the fitness of the current solution and the fitness of its neighbors. 

The other strategies combined with the observation functions $o^1$, $o^3$, $o^4$ obtain almost the same score, when tested on instances from the same distribution as used in the training phase. 

However, we will see in the next section that some of these strategies are more or less robust when tested on new instances of a different problem.

\subsection{Out-of-distribution tests \label{sec:outOfDistrib}}

Here, our aim is to check whether the best Neuro-LS strategies can also perform well on instances of different sizes and issued from distributions of instances on which they have not been trained.

We select the best  Neuro-LS strategies trained on the distribution of NK landscape instances with size $N=64$ and $K=8$,
and we test them on  new instances of different sizes\footnote{Note that a Neuro-LS strategy learned  for a given size of problem can be applied for instances of different sizes, as the input of the neural-network $g_{\theta}$ does not depend on $N$ (see Section \ref{sec:nn}).} of the Quadratic Unconstrained Binary Optimization (QUBO) problem.

The QUBO problem  is a single objective pseudo-Boolean optimization problem with quadratic interactions between binary variables. 
The objective function $f : \{0,1\}^N \to \mathbb{R}$ to maximize is defined by: 
$f(x) = x^{\intercal} Q x$
where $Q$ is a real matrix of dimension $N \times N$ and $x^{\intercal}$ is the transposed vector of $x$.
We consider a black-box optimization scenario with an unknown matrix $Q$. 

Moreover, we consider the \puboi~generator of QUBO instances \cite{TariVO22} able to bring QUBO with different properties. Indeed, the parameters of \puboi~generator can tune the density of matrix $Q$, as well as the importance of binary variables, and consequently the non-uniformity of the matrix. More formally, the fitness function is defined by 
$f(x) = \sum_{i=1}^{m} f_i(x_{i_1}, x_{i_2}, x_{i_3}, x_{i_4})$, where each sub-function $f_i$ is a quadratic function randomly selected from a set $\{ \varphi_1, \ldots, \varphi_4 \}$ where $\varphi_k$ has $2k$ symmetric local optima.\footnote{$\varphi_k$ are purely quadratic: $\forall x$, $\varphi_k(\overline{x}) = \varphi_k(x)$ where $\overline{x}$ is the complementary of $x$.} 
In \puboi, the binary variables are divided into two classes of importance: important, and non-important variables. For each sub-function $f_i$, the four variables $x_{i_j}$ are selected according to importance degree parameter $v$: the probability of selecting an important variable is proportional to the degree of importance. An additional parameter $\alpha$, called importance co-appearance, tuned the co-variance of selecting two important variables for the same sub-function $f_i$. See \cite{TariVO22} for more details. 

Table~\ref{tab:parameters} gives the experimental setup of \puboi~instances used in this work. We consider instances of size $N \in \{32,64,128,256\}$. The number of sub-functions $m$ tunes the density of the matrix ($16\%$ and $43\%$ for uniform instances respectively for the two values of $m$). Three types of interaction mechanisms are used between variables. The instances $I_{uni}$ have no specific important variables, \textit{i.e.} $I_{uni}$ instances are similar to QUBO problems with a random matrix. The instances $I_{imp}$ have important variables: the marginal probability of having important variables is equal to $v =10$ times the probability of non-important ones. Additionally, for the $I_{ic}$ instances, the value of the co-appearance parameter is high, the selection of important variables is not independent, and the selection of important variables is concentrated.
For each tuple of parameter values, we generate a test set $\mathcal{D}_{test}$ including $100$ independent instances.

\begin{table}[ht!]
    \caption{Parameters of \puboi~instances.}
    \label{tab:parameters}
    \centering
    \begin{small}
    \begin{tabular}{|c|c|c|}
        \hline
        Parameter & Description & Experimental values \\
        \hline
        $n$ & Problem dimension & $\{ 32, 64, 128 \} $\\ 
        $m$ & Number of sub-functions & $\{ 0.05, 0.2 \} \times \frac{n(n-1)}{2}$ \\ 
        $(v,\alpha)$ & (Degree of importance, & $I_{uni}: (1,1)$, $I_{imp}: (10,1)$,  \\
         &  co-appearance parameter) & $I_{ic}: (10, 1.09)$ \\ 
        \hline
    \end{tabular}
    \end{small}
\end{table}

Table \ref{tab:testQUBO} reports the average results obtained by the different strategies on the same test datasets $\mathcal{D}_{test}$. As in previous section, to insure a fair comparison for each instance, we use the same initial point to compute the trajectory produced by each LS strategy.

In Table \ref{tab:testQUBO}, the best average result obtained among the above-mentioned methods for each configuration of QUBO test instances is highlighted in bold. Results underlined indicate significant better results obtained by a Neuro-LS strategy on average compared to all the baseline strategies (p-value below 0.001), measured with a one-tailed Student t-test without assuming equal variance and with a Bonferroni correction to correct the significance threshold for multiple comparisons.

 This table shows that Neuro-LS strategies that use observation functions $o^3$ and $o^4$, which are based on ranking neighboring solutions, get the best results. These strategies outperform not only the baseline algorithms but also the Neuro-LS strategies that rely on raw information about neighbors' fitness (columns $o^1$ and $o^2$).

These two variants that use raw information on neighbors'  fitness perform very poorly, as they are not robust to changes in the amplitude of fitness values (as we will see in more detail in the next section). Indeed, these strategies were learned on distributions of instances of the NK problem with fitness values very different from those of the QUBO problem. 

In contrast, rank-based strategies are robust to changes in fitness function amplitudes. We also observe that these strategies perform well for all instance sizes, from 32 to 256, even though they were only trained on instances of size 64. These results suggest a degree of generality in the discovered strategies, indicating their potential effectiveness on larger instance sizes and novel problems.

When comparing results on columns $o^3$ and $o^4$, we observe that the introduction of additional z-score information in Neuro-LS produces better results for all instance types of larger size ($N=256$). As the choice of possible actions at each local search iteration becomes larger, using additional information on the fitness variations associated with each possible action seems beneficial. We will take a closer look at how the learned strategy exploits this information in the next section.

\begin{table}
\centering
\resizebox{0.95\textwidth}{!}{
\begin{tabular}{c c l|c|c|c|c|c|c|c}
\hline
\multicolumn{3}{c|}{Instances} & \multicolumn{7}{c}{Methods} \\ 
\cline{1-10} 
\multirow{2}{*}{$n$} & \multirow{2}{*}{$m$} & \multirow{2}{*}{$I$} & \multirow{2}{*}{\textbf{${\rm BHC}^{+}$}} & \multirow{2}{*}{\textbf{${\rm FHC}^{+}$}} & \multirow{2}{*}{\textbf{(1, $\lambda$)-ES}} & \multicolumn{4}{c}{\textbf{Neuro-LS}} \\
\cline{7-10}
 &  &  &  &  &  & $o^1$ & $o^2$ & $o^3$ & $o^4$  \\ \hline\hline

32 & $0.05$ & $I_{uni}$ & 58.7 & 59.0 & 58.7 & 1.1 & 12.4 &  60.6 & \textbf{61.7}\\ \hline
32 & $0.05$ & $I_{imp}$ & 44.4 & 42.4 & 43.2 & 2.7 & 10.9 &  \textbf{44.7} & 43.6\\ \hline
32 & $0.05$ & $I_{ic}$  & 41.7 & 41.5 & 41.3 & -0.18 & 5.3 &  41.1 & \textbf{41.7}\\ \hline
32 & $0.20$ & $I_{uni}$ & 131.4 & 133.5 & 131.4 & -0.75 & 85.1 &  139.6 & \underline{\textbf{144.3}}\\ \hline
32 & $0.20$ & $I_{imp}$ & 101.8 & 98.4 & 99.8 & -4.9 & 73.2 &  \textbf{103.8} & 100.7\\ \hline
32 & $0.20$ & $I_{ic}$ & 92.9 & 93.2 & 92.9  & -2.9  & 74.1 &  \textbf{93.7} & 92.0\\ \hline
\hline
64 & $0.05$ & $I_{uni}$ & 187.1 & 184.6 & 183.7 & -2.87 & 40.8 &  \underline{\textbf{195.4}} & \underline{\underline{194.4}}\\ \hline
64 & $0.05$ & $I_{imp}$ & 139.8 & 135.0 & 136.5 & -3.3 & 31.6 &  \underline{\textbf{141.1}} & 141.0\\ \hline
64 & $0.05$ & $I_{ic}$  & 131.1 & 129.4 & 130.4 & -0.85 & 25.9 &  \textbf{134.4} & 132.5\\ \hline
64 & $0.20$ & $I_{uni}$ & 396.8 & 391.3 & 393.8 &  6.36 & 133.9 & \underline{\textbf{418.2}} & \underline{415.9}\\ \hline
64 & $0.20$ & $I_{imp}$ & 310.7 & 305.1 & 312.6 & -2.78 & 201.4 &  \textbf{319.9} & 312.7 \\ \hline
64 & $0.20$ & $I_{ic}$  & 295.5 & 292.4 & 293.0 & 2.46 & 198.3 &  \textbf{303.9} & 301.1\\ \hline
\hline
128 & $0.05$ & $I_{uni}$ & 546.0 & 556.9 & 552.1 & 5.5 & 144.7 &  572.8 & \underline{\textbf{583.3}}\\ \hline
128 & $0.05$ & $I_{imp}$ & 423.1 & 414.0 & 418.9 & -8.3 &  75.3 & \textbf{432.5} & 432.4 \\ \hline
128 & $0.05$ & $I_{ic}$  & 406.0 & 402.0 & 405.1 & 3.2 & 132.1 &  420.4 & \underline{\textbf{424.1}}\\ \hline
128 & $0.20$ & $I_{uni}$ & 1156.7 & 1171.4 & 1123.0 & -13.4 & 338.1&  \underline{1215.1} & \underline{\textbf{1228.0}}\\ \hline
128 & $0.20$ & $I_{imp}$ & 901.6 & 899.1 & 895.6 & -8.97 & 256.5 & 923.3 & \textbf{919.9}\\ \hline
128 & $0.20$ & $I_{ic}$  & 888.3 & 893.7 & 881.1 & 11.7 & 399.4 &  \textbf{917.1} & 914.0\\ \hline
\hline
256 & $0.05$ & $I_{uni}$ & 1620.7 & 1630.6 & 1609.5 & 7.8 &  438.8 & \underline{1681.3} & \underline{\textbf{1701.0}}\\ \hline
256 & $0.05$ & $I_{imp}$ & 1275.0 & 1252.6 & 1275.8 & -4.12 & 342.4 &  1299.5 & \underline{\textbf{1320.9}} \\ \hline
256 & $0.05$ & $I_{ic}$  & 1226.6 & 1230.6 & 1235.3 & 1.38& 262.1 &  \underline{1276.6} & \underline{\textbf{1295.1}}\\ \hline
256 & $0.20$ & $I_{uni}$ & 3333.6 & 3376.1  & 3282.9 & -13.3 & 1083.8 &  \underline{\textbf{3485.0}} & \underline{3484.6} \\ \hline
256 & $0.20$ & $I_{imp}$ & 2604.0 & 2580.3 & 2574.3 & -20.4 & 693.9 &  \underline{2687.0} & \underline{\textbf{2697.6}}\\ \hline
256 & $0.20$ & $I_{ic}$  & 2592.9  & 2574.4 & 2584.4 & 8.3 & 868.5 &  \underline{2698.6} & \underline{\textbf{2713.4}}\\ \hline
\end{tabular}
}
\caption{Average score (fitness value) on test instances obtained by the baseline local search strategies
and the Neuro-LS strategies for different QUBO problems of different sizes. Best result on each line is written in bold. Results underlined indicate significant better results obtained by a Neuro-LS strategy on average compared to all the baseline strategies (p-value below 0.001), measured with a one-tailed Student t-test without assuming equal variance and with a Bonferroni correction to correct the significance threshold for multiple comparisons. }
\label{tab:testQUBO}
\end{table}

\subsection{Study of Neuro-LS emerging strategies \label{sec:emerging}}


The objective here is to analyse in detail the best Neuro-LS strategies that have emerged with neuro-evolution and to understand their decision-making processes for the different types of NK landscapes on which they have been trained (smooth or rugged) and for the different types of information given as input to the neural network.

\subsubsection{Emerging strategies for smooth landscape}

For smooth NK landscapes (when $K=1$ or $K=2$) and for all the observation functions used as input, Neuro-LS almost always learns to perform  a best improvement move, which explains why for these instances, it obtains almost the same score as the ${\rm BHC}^{+}$ strategy (see Table~\ref{tab:test}).

\begin{figure}[h!]
\includegraphics[width=10cm]{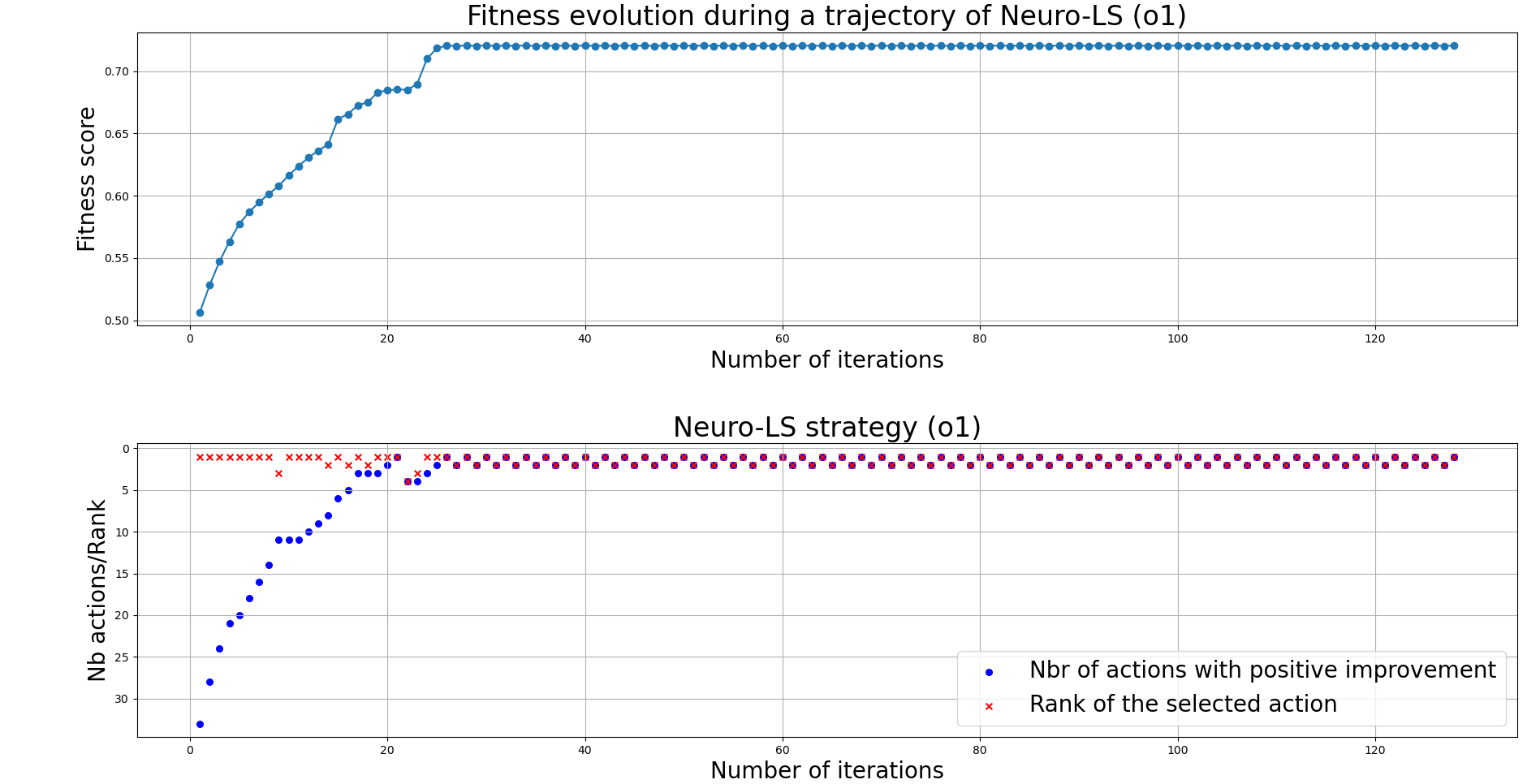}
\caption{Fitness evolution curve and strategy used by Neuro-LS using $o^1$ observations for the resolution of an instance with smooth NK landscape ($N=64$ and $K=2$). \label{fig:emerging_strat_smooth}} 
\end{figure}

Figure \ref{fig:emerging_strat_smooth} displays a representative example of the trajectory performed by Neuro-LS with $N=64$ and $K=2$ and when using the observation vector $o^1(x)$ as input (vector of delta fitness). 

This figure shows two graphs based on data collected during the resolution of this instance. The graph on top of this figure shows the evolution of the fitness reached by Neuro-LS  over the $H = 2 \times N = 128$ iterations.  The graph below shows at each iteration  the number of available actions  corresponding to an improvement of the score (in blue), and  the rank of the action selected by  Neuro-LS, measured in terms of fitness improvement (in red). A rank of 1 on this plot indicates that Neuro-LS has chosen a best improvement move, while a rank of 64 indicates a worst deteriorating move (note that the y-axis is inverted, because it is a maximization problem).

First of all, we can see from this graph that at the start of the search, there are around 32 score-improving actions (blue dot on the left and on the lower graph), which unsurprisingly corresponds for an instance of size 64 to as many score-improving moves as score-reducing moves among the neighbors of a randomly chosen point in the search space.

Neuro-LS almost always chooses the best improving move (red dots of rank 1 on the lower graph). However we note that it  chooses sometimes an action that corresponds to a rank between 2 and 4. It occurs when fitness becomes high to avoid climbing in certain cases towards the first local maximum encountered. This choice may explain why Neuro-LS sometimes achieves slightly better results than the ${\rm BHC}^{+}$ strategy (see Table~\ref{tab:test}), although it starts from the same starting points for all test instances (note that  all these strategies are deterministic).

After around 25 iterations, Neuro-LS has reached a local maximum and cannot escape from it. For this type of instance, with $K=1$ and $K=2$, the landscape is indeed smoother, but it is made up of large basins of attraction that are difficult to escape without using a quite sophisticated perturbation mechanism, which seems impossible to learn in our context (when only one-flip moves can be selected and the strategy is memoryless).

\subsubsection{Emerging strategies for rugged landscape \label{sec:emerging_rugged}}

 For rugged landscapes, when $K=4$ and $K=8$, the emerging strategy is much more interesting. 
 We detail here the most interpretable strategy learned using observation vectors $o^1$ and $o^3$, respectively corresponding to  variations of fitness for each possible move  and a ranking of the improving and deteriorating moves (see Section \ref{sec:observations}). The two other strategies using $o^2$ and $o^4$ observation functions are described in Appendix \ref{appendix}.

 \paragraph{Emerging strategy based on the variations of fitness for each possible move}

\begin{figure}[h!]

\includegraphics[width=10cm]{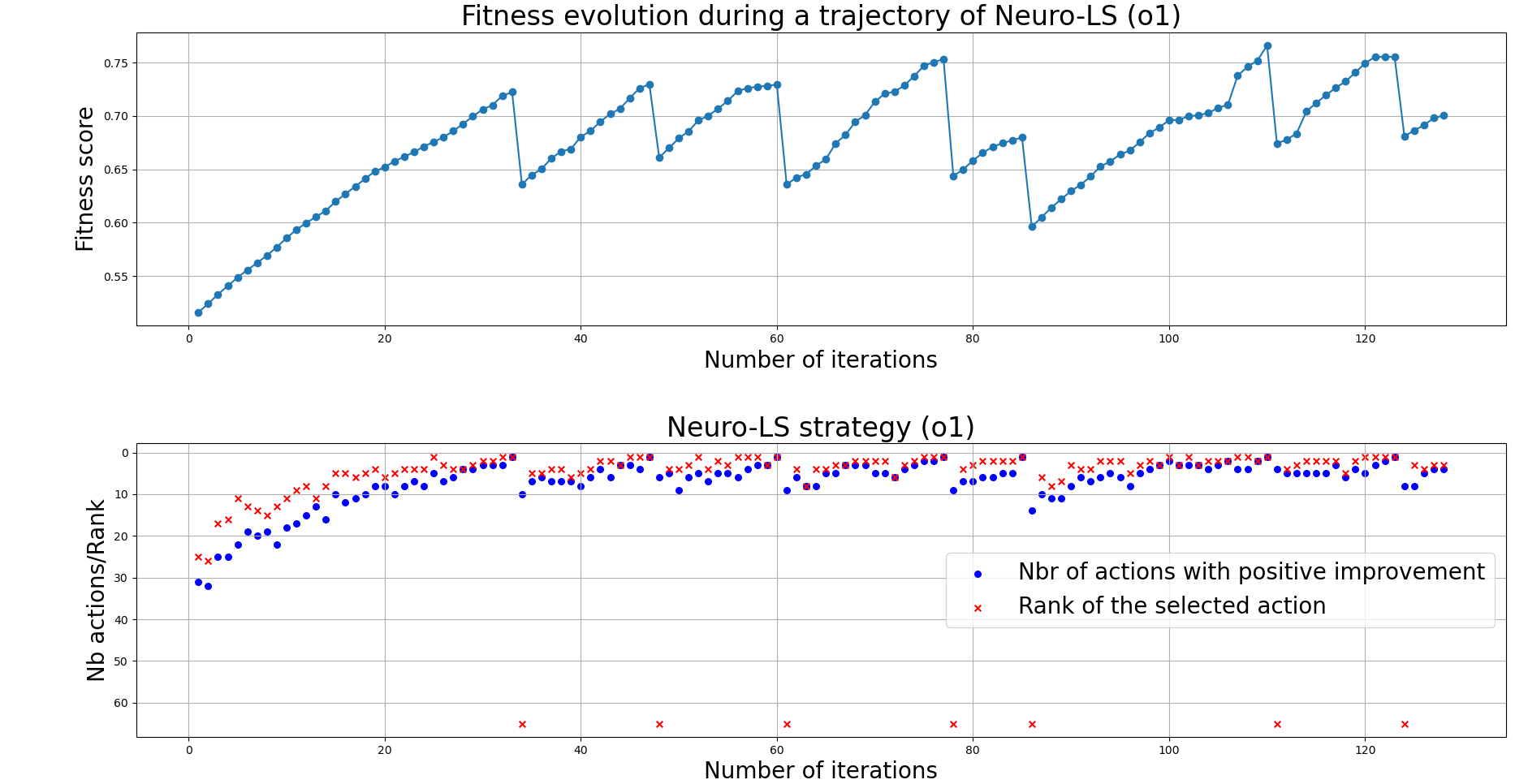}
\caption{Fitness evolution curve and strategy used by Neuro-LS using $o^3$ observations for the resolution of an instance with significantly rugged NK landscape ($N=64$ and $K=8$). \label{fig:emerging_strat_o1}} 
\end{figure}

 Figure \ref{fig:emerging_strat_o1} displays a representative example of the trajectory performed by Neuro-LS using observation $o^1$ when $N=64$ and $K=8$. On this plot, we observe that the emerging Neuro-LS strategy has two successive operating modes:

\begin{enumerate}
        \item \textbf{Small steps hill climbing behavior.} When the number of actions corresponding to positive improvement of the score, $N_a^+$ is greater than 0, Neuro-LS does not always choose the best improvement flip, but rather a flip associated with a small increase in fitness (as can be seen in the top graph). This avoids being trapped too quickly in a local optimum.
        
        \item \textbf{Jump with worst move.} When there is no more improving move, Neuro-LS does not stagnate, but instead directly chooses to perform the worst possible move (with rank 64). Even if this movement considerably deteriorates the current fitness score, 
        it actually maximizes its long-term chances of escaping the current local optimum and continually exploring new areas of the search space. Indeed, we observe on this plot that Neuro-LS continuously improves its score with this strategy for this instance. Note that after choosing the worst possible move, it does not choose the best possible move, otherwise it would return to the same local optimum.  
    \end{enumerate}

\begin{figure}[h!]
\centering
\includegraphics[width=10cm]{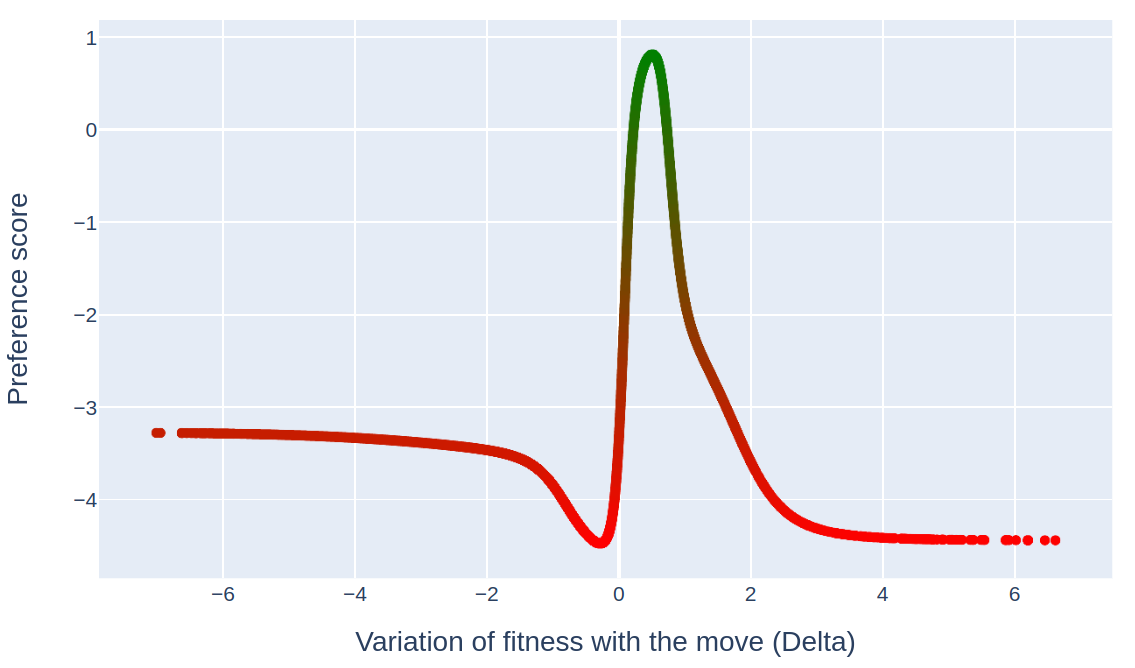}
\caption{Output of the neural network $g_{\theta}(\Delta_i(x)))$ (y-axis) for each input $\Delta_i(x)$ (x-axis). It corresponds to the values collected during 10 trajectories of Neuro-LS on the test instances of the NK Landscape problem with $N=64$ and $K=8$.   \label{fig:function_strat_o1}} 
\end{figure}

To explain why Neuro-LS has this behaviour, we display in Figure  \ref{fig:function_strat_o1} the continuous function $g_{\theta}$ learned by the neural network. It explains why Neuro-LS chooses one movement over another. Indeed, as seen in Section \ref{sec:nn}, given the  vector of observation, which in this case is equal to $o^1(x) = (\Delta_1(x), \dots, \Delta_N(x))$, the neural network outputs a vector $g_{\theta}(o) = (g_{\theta}(\Delta_1(x)), \dots, g_{\theta}(\Delta_N(x)) \in \mathbb{R}^{N}$. Then, the action $a = \argmax_{i\in \llbracket 1,N \rrbracket} g_{\theta}(o_i)$ is chosen by Neuro-LS. Thus, the output of the neural network, $g_{\theta}(\Delta_i(x))$, can be interpreted as a preference score for a move associated to a variation of fitness $\Delta_i(x)$.

As seen in Figure \ref{fig:function_strat_o1}, the preference score is greater for moves that corresponds to a small variation of fitness (in green), which explains the first operating mode mentioned earlier, with Neuro-LS performing only small steps and avoiding large improvement of fitness, which may cause it to get trapped too early in a local maximum.
When there are only movements with negative score improvements, we see that $g_{\theta}$ gives a higher preference score for the most deteriorating move, which explains the second operating mode, when Neuro-LS performs a jump, when there are no more score-improving moves.

Note that this strategy depends on the amplitude of the variations of fitness of the problem. If the fitness function undergoes a change of scale, the strategy learned by Neuro-LS is completely disrupted. This explains why the same Neuro-LS strategy using the $o^1$ observation vector does not score well when applied to the UBQP pseudo-Boolean problem, which does not have the same fitness variation amplitudes (see Table \ref{tab:testQUBO}). We will see in the following that using an information based on the ranked allows to avoid this problem.

\paragraph{Emerging strategy based on a ranking of improving and deteriorating moves}

\begin{figure}[h!]

\includegraphics[width=10cm]{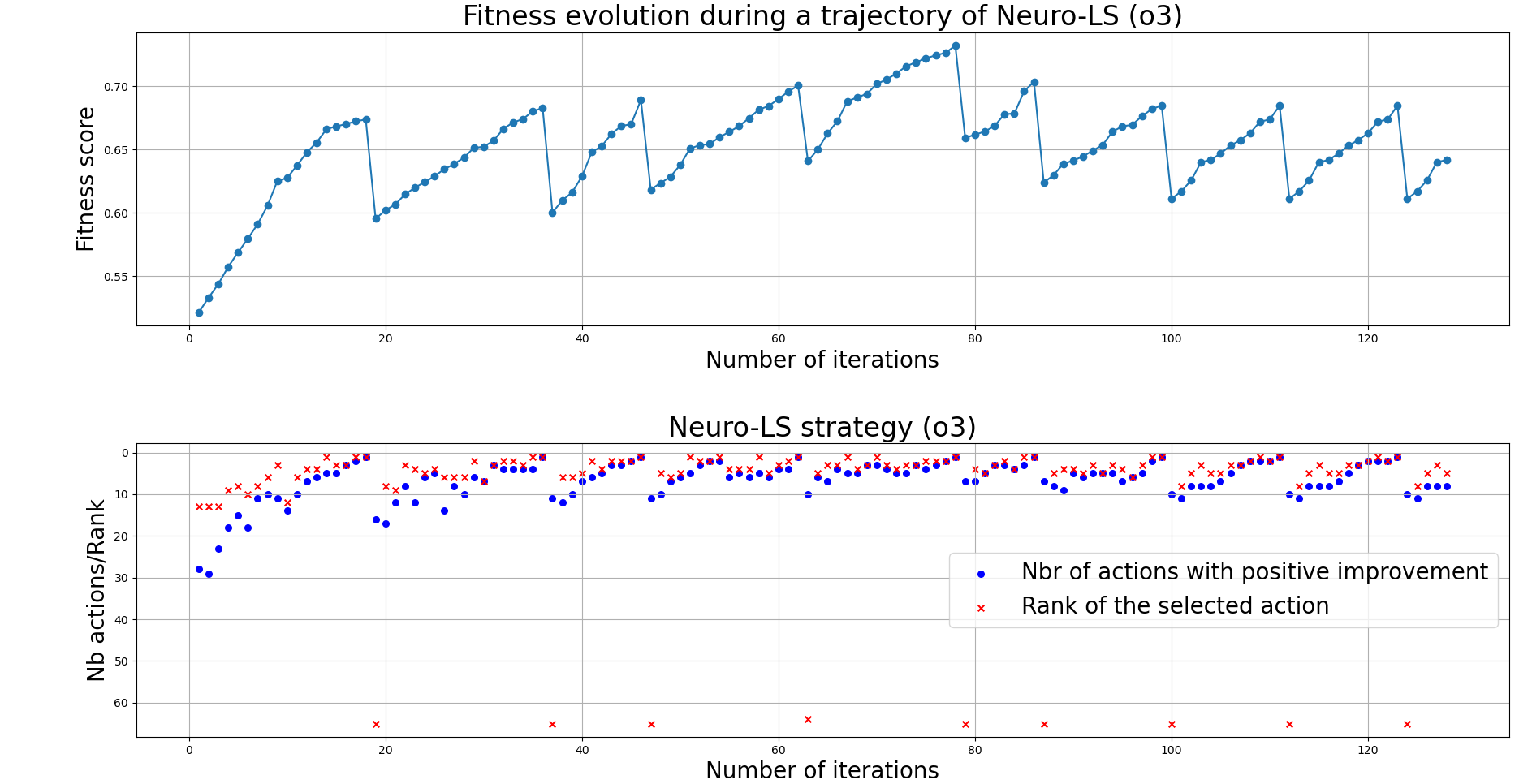}
\caption{Fitness evolution curve and strategy used by Neuro-LS using $o^3$ observations for the resolution of an instance with significantly rugged NK landscape ($N=64$ and $K=8$). \label{fig:emerging_strat_o3}} 
\end{figure}

Figure \ref{fig:emerging_strat_o3} displays a representative example of the trajectory performed by Neuro-LS using observation $o^3$ when $N=64$ and $K=8$. 

We observe on this plot that the emerging Neuro-LS strategy has two successive operating modes:

\begin{enumerate}
        \item \textbf{Median hill climbing behavior.} When the number of actions associated with a positive improvement of the score, $N_a^+$ is greater than 0, Neuro-LS does not always choose the best improvement movement,  but instead a move with a rank approximately equal to $N_a^+/2$. This provides a good compromise between improving the score and avoiding being trapped too quickly in a local optimum.  
        \item \textbf{Jump with worst move.} When there is no more improving move, Neuro-LS does not stagnate, but instead directly also chooses to perform the worst possible move (with rank 64). 
    \end{enumerate}

\begin{figure}[h!]
\centering
\includegraphics[width=10cm]{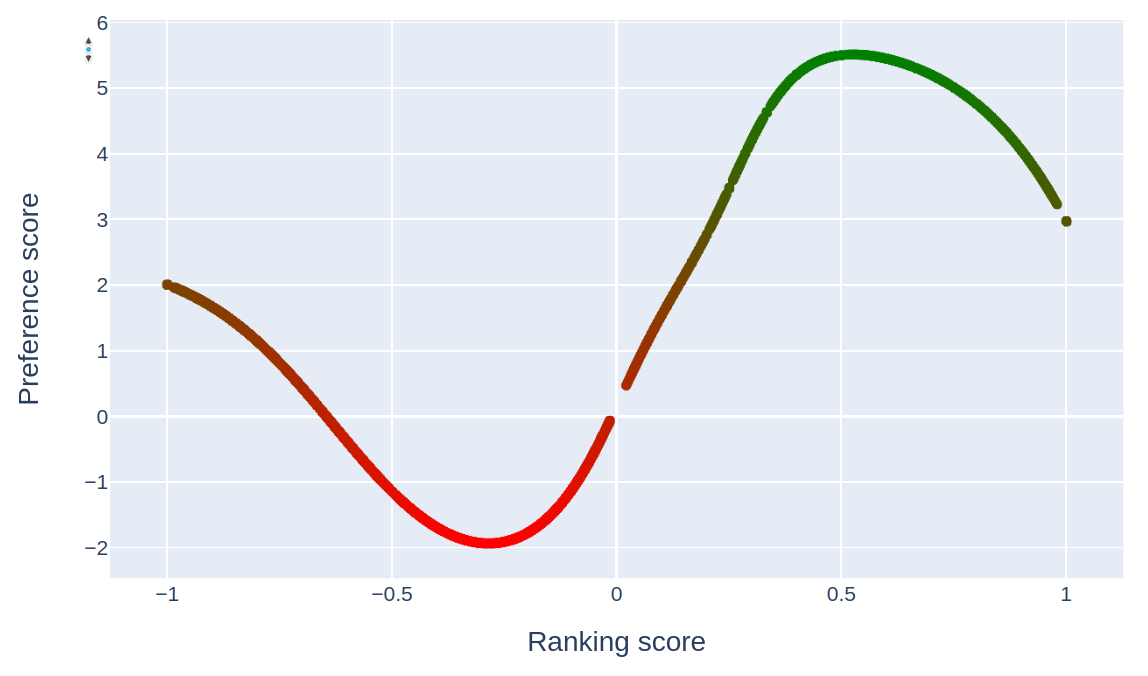}
\caption{Output of the neural network $g_{\theta}(o^3(x)_i))$ (y-axis) for each input $o^3(x)_i$ (x-axis). It corresponds to the values collected during 10 trajectories of Neuro-LS on the test instances of the NK Landscape problem with $N=64$ and $K=8$.  ). \label{fig:function_strat_o3}} 
\end{figure}

To explain why Neuro-LS has this behaviour, we display in Figure  \ref{fig:function_strat_o3} the continuous function $g_{\theta}$ learned by the neural network. We remain the reader that the values $o^3(x)_i$ given as input to the neural network corresponds to a ranking of the improvement moves, which receive a strictly positive value in the range $]0,1]$, and the deteriorating moves, which receive a strictly negative value in the range $[-1,0[$. We see on this figure that Neuro-LS gives the higher preference score for a median improving move (with a value around 0.5), which explain the median hill climbing behavior observed during the trajectory. When there is only negative moves, Neuro-LS prefers the worst deteriorating move with a value of -1, which explain the jump  when there are no more improving moves.

Note that for the NK landscape instances we are considering, there is almost never any move with strictly zero variation in fitness, which explains why we have not collected any entries with a value of 0, and therefore why we observe a discontinuity on this curve around the value of 0 (but $g_{\theta}$ is indeed a continuous function in this case).

Note that we have also tested this emerging strategy learned on rugged landscape (when $K=4$ or $K=8$) on instances generated with smooth landscape ($K=1$ or $K=2$). It does not work very well as the basins of attraction are too large when $K=1$ or $K=2$, making the jump ineffective in this case. This is why, when $K=1$ or $K=2$, Neuro-LS learns a completely different strategy consisting of almost always choosing a better improvement move at each iteration in order to rapidly converge towards a local optimum.

\color{black}

\section*{Conclusion}

Our study explores the emergence of new local search algorithms with neuro-evolution. Results on NK landscapes show that different neural network policies are learned, each adapted to the resolution of a particular landscape distribution type (smooth or rugged). Our algorithm is competitive with basic deterministic local search procedures for all the NK landscape types considered in this work. 
Particularly for rugged landscapes, it can achieve significantly better results with an original emerging strategy, using a worst-case improvement move, which proves very effective in the long run for escaping local optima. We have also shown in this work that emergent strategies based on the rank of improvement and deterioration moves are very robust. They can easily be used for instances of different sizes, but also for  other pseudo-Boolean problems (as we saw with the out-of-distribution tests on instances of the QUBO problem).

This study outlines avenues for future research on the automatic discovery of more advanced strategies using as input a richer set of observations to make its decision. The proposed framework could also be applied to study the emergence of strategies adapted to other types of combinatorial optimization problems.
  
\section*{Acknowledgment}

We would like to thank Pr. S\'ebastien Verel for his suggestion to use a notion of rank as input to the neural network in order to improve the robustness of the approach, and also for providing us with the UBQP instances used in this paper.

This work was granted access to the HPC resources of IDRIS (Grant No. AD010611887R1) from GENCI.



\section*{Funding information}

The authors would like to thank the Pays de la Loire region for its financial support for the Deep Meta project (Etoiles Montantes en Pays de la Loire). The authors also acknowledge ANR – FRANCE (French National Research Agency) for its financial support of the COMBO project (PRC - AAPG 2023 - Axe E.2 - CE23).  









\bibliography{biblio}

\begin{appendix}

\section{Other emerging strategies \label{appendix}}

In this section we analyze the other strategies that emerged when the $o^2$ and $o^4$ observation functions were used. Analyses of these strategies are a little less easy to interpret, as in this case the neural network learned is a non-linear function of two input variables instead of one. 

\paragraph{Emerging strategy based on raw information about current and neighboring fitness}

\begin{figure}[h!]
\centering
\includegraphics[width=11cm]{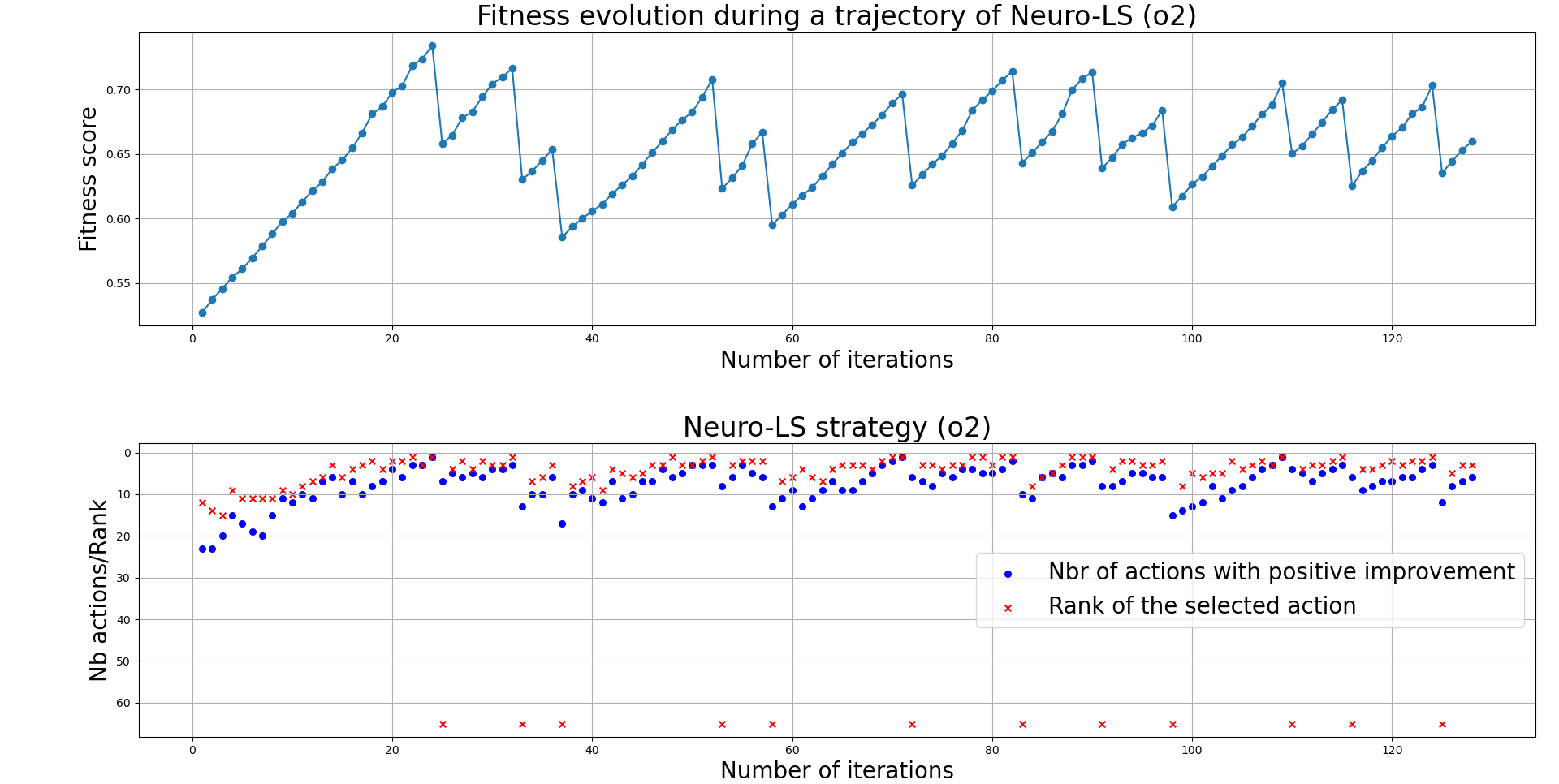}
\caption{Fitness evolution curve and strategy used by Neuro-LS using $o^2$ observations for the resolution of a instance with significantly rugged NK landscape ($N=64$ and $K=8$). \label{fig:emerging_strat_o2}} 
\end{figure}

Figure \ref{fig:emerging_strat_o2} displays a representative example of the trajectory performed by 
Neuro-LS using observation $o^2$ when $N = 64$ and $K = 8$. We observe on this 
plot that the emerging Neuro-LS strategy has two successive operating modes like the   strategy based on the variations of fitness for each
possible move (see Section \ref{sec:emerging_rugged}): small steps hill climbing behavior and Jump with worst move.
It is interesting to see that this same behavior can also emerge when using very raw information as input.

However we see in this plot that the strategy is slightly less imprecise to find the local maxima, because we see in Figure \ref{fig:emerging_strat_o2} that the jump with worst move can be triggered even if there are still improving moves. As a result, this strategy can sometimes miss opportunities to find a good solution, which explain why this strategy obtain less good results as reported in Table \ref{tab:test}.

\begin{figure}[h!]
\centering
\includegraphics[width=11cm]{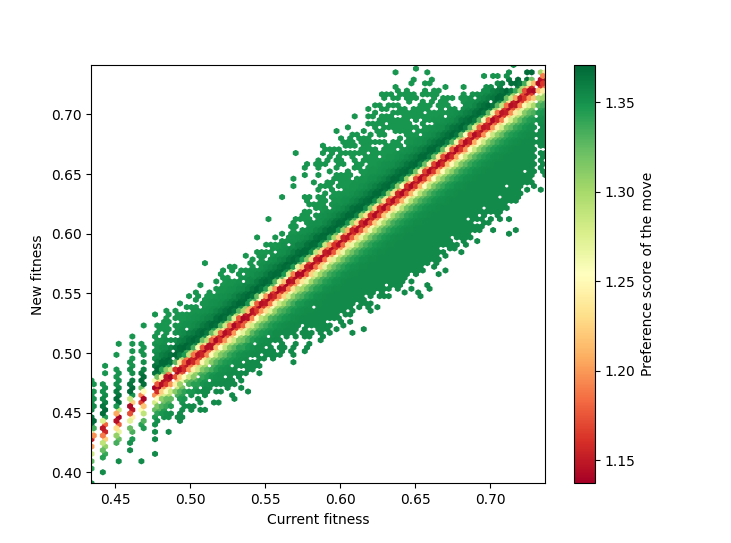}
\caption{Output of the neural network $g_{\theta}(o^2(x)_i)$, or preference score for the move $i$, for each 2-dimensional input with the current fitness $f_{{\rm obj}}(x)$ (x-axis) and the new fitness $f_{{\rm obj}}(flip_i(x))$ (y-axis). It corresponds to the values collected during 10 trajectories of Neuro-LS on the test instances of the NK Landscape problem with $N=64$ and $K=8$. \label{fig:function_strat_o2}} 
\end{figure}

We display in Figure  \ref{fig:function_strat_o2} the continuous function $g_{\theta}$ learned by the neural network taken a 2-dimensional vector as input. We see in this figure that the Neuro-LS strategy also gives the worst preference in this case to moves that deteriorate slightly, when the new fitness (after the move) is slightly lower than the current fitness (red hexagons on the heatmap). We see in dark green that the most preferred moves are the small improvement moves. The  strong deterioration moves have also a quite good preference score, when the new fitness after the flip is really inferior to the current fitness (allowing for jumping moves).

\paragraph{Emerging strategy based on ranking and z-score}

\begin{figure}[h!]
\centering
\includegraphics[width=11cm]{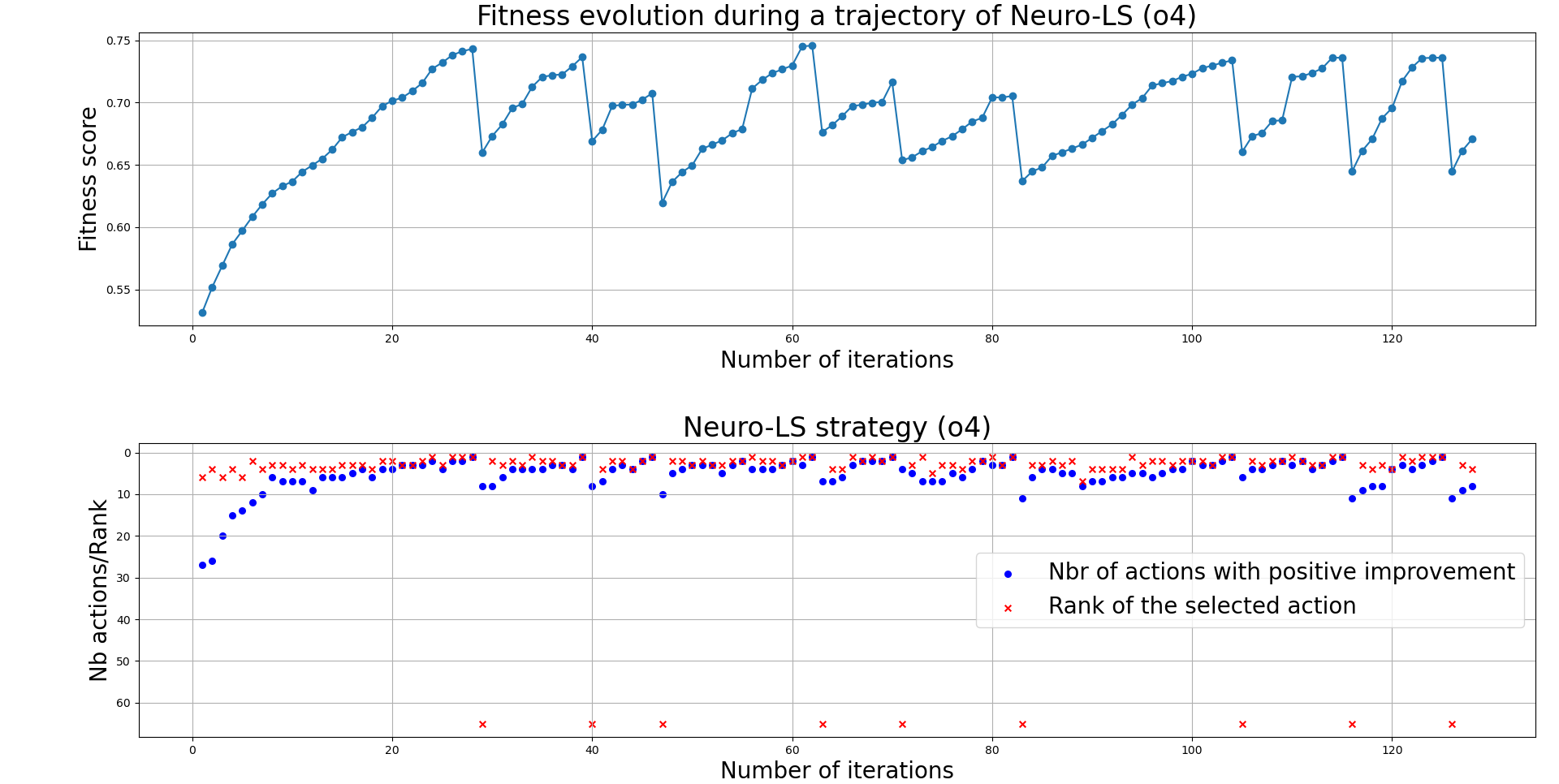}
\caption{Fitness evolution curve and strategy used by Neuro-LS using $o^4$ observations for the resolution of a instance with significantly rugged NK landscape ($N=64$ and $K=8$). \label{fig:emerging_strat_o4}} 
\end{figure}

Figure \ref{fig:emerging_strat_o4} displays a representative example of the trajectory performed by 
Neuro-LS using observation $o^4$ when $N = 64$ and $K = 8$. We observe on this 
plot that the emerging Neuro-LS strategy has also two successive operating modes like the  other strategies: small steps hill climbing behavior and jump with worst move.

However, we can see from this figure that the strategy still allows itself to choose moves that are among the most improving (on the left of the bottom graph), even if they are moves that improve fitness very little. We can see that before the jump, this strategy had already achieved a very good score of almost 0.75, much better for the same instance and starting point than the other strategies presented above.

\begin{figure}[h!]
\centering
\includegraphics[width=11cm]{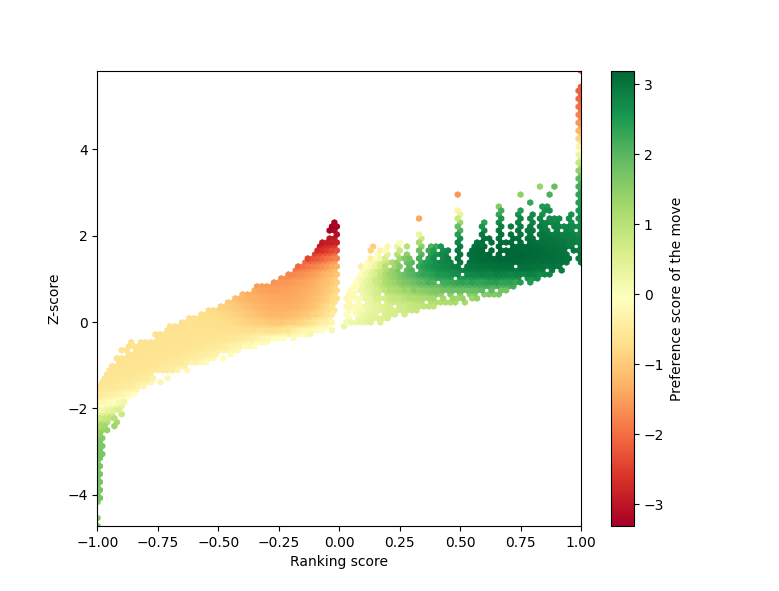}
\caption{Output of the neural network $g_{\theta}(o^4(x)_i)$, or preference score for the move $i$, for each 2-dimensional input with the ranking score $o^3(x)_i$ (x-axis) and the z-score $Z(\Delta_i(x))$ (y-axis). It corresponds to the values collected during 10 trajectories of Neuro-LS on the test instances of the NK Landscape problem with $N=64$ and $K=8$. \label{fig:function_strat_o4}} 
\end{figure}

This is due to a clever combination of the rank of the improving movements and their z-score, as shown in figure \ref{fig:function_strat_o4}. Indeed, we can see on this figure that the points preferred in green can correspond to points whose ranks are situated between the median move and the best move, but always controlled so as not to have too high a z-score (i.e. so as not to choose a move that would improve the score too much in comparison with the other possible moves).

On the right-hand side of this figure, we can see that movements with a very high z-score (which also corresponds to the best improving moves) are the most penalized, as these are the ones that most quickly lead to being stuck in a local maximum.

On the left of this figure, we can also see that very deteriorating moves with a very negative z-score also have a fairly favorable score, as they are the ones that allow Neuro-LS to make a jump when there are no more improving moves.

\end{appendix}
\end{document}